\newif\ifarxiv
\newif\ifralfinal
 \ralfinalfalse \arxivtrue

\ifarxiv
\documentclass[letterpaper, 10 pt, conference]{ieeeconf}  %
\fi
\ifralfinal
\documentclass[letterpaper, 10 pt, journal, twoside]{IEEEtran}
\fi

\IEEEoverridecommandlockouts                              %

\usepackage{graphics} %
\usepackage{epsfig} %
\usepackage{mathptmx} %
\usepackage{times} %
\usepackage{amsmath} %
\usepackage{amssymb}  %
\usepackage{multirow}
\usepackage{url}
\usepackage{comment}
\usepackage[dvipsnames]{xcolor}
\usepackage{booktabs}
\usepackage{wrapfig}
\usepackage{balance}
\usepackage{comment}
\ifarxiv
\usepackage{pdfpages}
\usepackage{fancyhdr}
\fi

\begin{document}

\title{\ifarxiv\LARGE \bf\fi

MultiRes-NetVLAD: Augmenting Place Recognition Training with Low-Resolution Imagery 

}

\author{Ahmad Khaliq, Michael Milford and Sourav Garg%
\thanks{All Authors are with the School of Electrical Engineering and Robotics and QUT Centre of Robotics at
Queensland University of Technology (QUT), Brisbane, Australia}
\thanks{\tt\small Email: ahmad.khaliq@hdr.qut.edu.au }%
}

\maketitle
\ifarxiv
\thispagestyle{fancy}
\pagestyle{plain}
\fi

\begin{abstract}

Visual Place Recognition (VPR) is a crucial component of 6-DoF localization, visual SLAM and structure-from-motion pipelines, tasked to generate an initial list of place match hypotheses by matching global place descriptors. However, commonly-used CNN-based methods either process multiple image resolutions \textit{after} training or use a \textit{single} resolution and limit multi-scale feature extraction to the last convolutional layer during training. In this paper, we augment NetVLAD representation learning with low-resolution image pyramid encoding which leads to richer place representations. The resultant multi-resolution feature pyramid can be conveniently aggregated through VLAD into a single compact representation, avoiding the need for concatenation or summation of multiple patches in recent multi-scale approaches. Furthermore, we show that the underlying learnt feature tensor can be combined with existing multi-scale approaches to improve their baseline performance. Evaluation on 15 viewpoint-varying and viewpoint-consistent benchmarking datasets confirm that the proposed \textit{MultiRes-NetVLAD} leads to state-of-the-art Recall@N performance for global descriptor based retrieval, compared against 11 existing techniques. Source code is publicly available at \url{https://github.com/Ahmedest61/MultiRes-NetVLAD}.

\end{abstract}

\section{INTRODUCTION}

\ifralfinal
\else
Have \fi I seen this place before? – understanding and recognizing a revisited place in a pre-built environment map~\cite{lowry2015visual} has been of great interest to researchers for the last two decades. This problem is often addressed as Visual Place Recognition (VPR)~\cite{garg2021your,masone2021survey}. 
VPR is the cornerstone of many robotic tasks and applications such as Simultaneous Localization and Mapping (SLAM)~\cite{cadena2016past} for image sequences, Structure-from-Motion (SfM)~\cite{Sarlin2021BackTT,schonberger2016structure} for unordered sets of images and 6-DoF localization~\cite{zhang2021visual,gabriele2021master} for given prior maps.
VPR is challenging due to uncontrollable environmental factors such as illumination, viewpoint, and seasonal transitions~\cite{lowry2015visual,garg2021your,doi:10.1177/0278364916679498}.
Therefore, learning a generic but robust place representation has become an active area of research.

\begin{figure}
	\centering
	\includegraphics[trim={0 3cm 0 3cm},clip,width=0.5\textwidth]{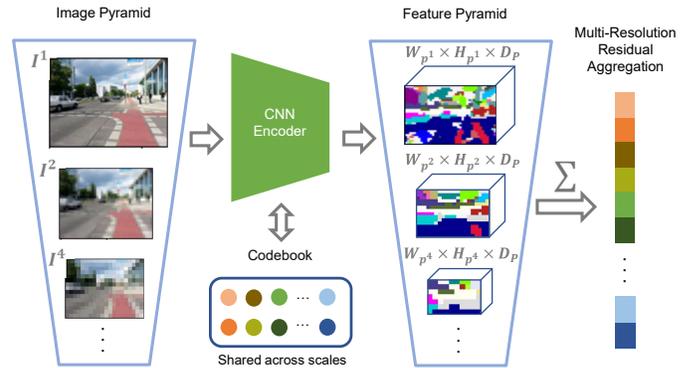}
	\caption{We augment NetVLAD representation learning with low-resolution image pyramid, aggregating a union of features across multiple resolutions to obtain a compact, but more representative, global place descriptor.}
	\label{figure:frontPage}
\end{figure}

Existing VPR research has delivered a variety of robust techniques such as global image descriptor learning~\cite{arandjelovic2016netvlad,chen2018learning}, local feature matching~\cite{hausler2021patch}, GANs-based image translation~\cite{8461081}, and use of additional information in the form of semantics~\cite{garg2018lost,7989305,DBLP:journals/ftrob/GargSDMCCWCRGCM20}, depth~\cite{6594910}, odometry~\cite{pepperell2014all}, point clouds~\cite{7989618,8593562}, and sequences~\cite{milford2012seqslam,garg2021seqmatchnet,Garg2021SeqNetLD,neubert2021vector,schubert2021fast}.

Recent studies have focused more on improving the robustness of single-image based place description by inducing \textit{multi-scale} information within the global descriptors~\cite{8700608,8382272,le2020city,8961602,xin2019localizing,mao2019learning}. However, these CNN-based methods either process multiple image resolutions \textit{after} training~\cite{8961602,xin2019localizing,mao2019learning} or use a \textit{single} resolution and limit multi-scale feature extraction to the last convolutional layer during training~\cite{8700608,le2020city}. These choices limit the full utilization of learnable features that exist at different resolutions of the original image, especially at lower fidelities, which are not necessarily captured by typical CNNs despite their pyramid structure. 

In this paper, we augment NetVLAD representation learning with low-resolution image pyramid encoding to generate richer and more performant place representations, as shown in Figure~\ref{figure:frontPage}. Our main contributions are as below:
\begin{itemize}
    \item a novel Multi-Resolution feature residual aggregation method, dubbed MultiRes-NetVLAD, based on a shared multi-resolution feature vocabulary;
    \item a resultant image encoder amenable to different feature aggregation strategies and applicable to both viewpoint-consistent and viewpoint-varying datasets, leading to state-of-the-art recall performance; and
    \item analyses that demonstrate i) increased test-time robustness to variations in image resolutions, ii) the effect of coverage and density of multi-resolution configurations, and iii) ablations of training strategies with varying use of low-resolution imagery. %
\end{itemize}

The remainder of this paper is organized as follows: Section \ref{section:lit_review} reviews the relevant multi-scale CNN-based place recognition works; Section \ref{section:prop_method} describes the proposed multi-resolution feature learning method; Section \ref{section:imp_details} discusses implementation details and benchmark datasets and methods; Section \ref{section:results} presents the experimental results and analyses; and Section \ref{section:conclus} presents conclusions with future work directions.

\section{LITERATURE REVIEW}
\label{section:lit_review}

Visual place recognition in complex urban areas is challenging due to uncertain environmental variations such as illumination~\cite{6594910}, seasons~\cite{sunderhauf2013we}, viewpoints~\cite{chen2017deep,chen2018learning}, dynamic instances~\cite{zaffar2020memorable} or structural changes~\cite{zaffar2020vpr}. Based on the Bag-of-Words (BoW)~\cite{filliat2007visual} models, earlier works~\cite{cummins2011appearance} focused on identifying unique image regions using handcrafted local (SIFT~\cite{lowe2004distinctive}, SURF~\cite{bay2006surf}, CoHoG \cite{8972582}) or global (HoG~\cite{dalal2005histograms}, Gist\cite{oliva2006building}) feature detectors. %
More recently, fisher vector (FV)~\cite{babenko2015aggregating} and VLAD~\cite{jegou2010aggregating} have emerged as powerful alternatives to the BoW scheme. 
However, handcrafted methods still lack robustness which can potentially be achieved through data-driven learning-based approaches.

\subsection{Deep Learning and Global Descriptors} With the prevalence of deep learning in vision-centric tasks including VPR, as reviewed in~\cite{masone2021survey,zhang2021visual,garg2021your,7301270}, several works have focused on CNN-based global feature pooling techniques, for example, R-MAC~\cite{Tolias2015ParticularOR}, SPoC~\cite{chen2017only}, HybridNet~\cite{chen2017deep}, NetVLAD~\cite{arandjelovic2016netvlad}, GeM~\cite{8382272}), LoST~\cite{garg2018lost}, DeLG~\cite{cao2020unifying}
and AP-GeM~\cite{revaud2019learning}. Such frameworks introduce local region-level re-weighting that assists in generating a more suitable image representation for performing place recognition under challenging camera viewpoint and appearance variations~\cite{zaffar2019levelling}. Authors in \cite{gordo2017end} train the R-MAC (max-pool across several multi-scale overlapping regions) framework on Triplet loss-based ranking, and learn compact global feature representations for instance-level image retrieval. Global compact descriptors based VPR remains a go-to solution for fast retrieval of place match hypotheses~\cite{garg2021your}, and is also the motivation of this work.

\subsection{Leveraging Multi-scale Information}
Several studies~\cite{chen2017only,hausler2021patch,xin2019localizing,8961602,8700608,le2020city} have demonstrated that incorporating spatial information, especially from within the convolutional layers, for example, pyramid patches~\cite{8700608} or regions~\cite{chen2016attention,khaliq2018holistic} can significantly improve place recognition performance. These methods differ in terms of whether or not multi-scale information is part of the training process, and how exactly more spatial context is gathered when considering multiple scales, for example, using different image resolutions versus using different patch/region sizes, as reviewed below.

\subsubsection{Multi-Scale Processing `After' Training}
Since the use of multi-scale information within the training process is not always trivial, many researchers have proposed multi-scale techniques that post-process CNN features or use multiple image resolutions. \cite{hausler2021patch} post-processed NetVLAD's last convolutional layer with multiple patch sizes to improve local feature matching. \cite{8382272} used multiple image resolutions and then aggregated the final feature maps to obtain a compact representation. \cite{Xin2019RealTimeVP} used multi-scale CNN landmarks analysis to select robust features for VPR. However, post-processing CNN features at multiple scales may not fully capture the features that are directly relevant for VPR.

\subsubsection{Learning with Multi-Scale}
Recent works~\cite{8961602,xin2019localizing} have attempted to improve VPR performance by incorporating multi-scale features in the training process. In~\cite{8961602}, authors propose a trainable end-to-end framework with a deep fusion of multi-layer max-pooled convolutional features, achieved through the use of convolutional kernels of varying sizes.
\cite{Zhu_2018} learns pyramid attention pooling using a spatial grid defined on top of last convolutional layer which is finally summed to obtain a compact descriptor. Based on a multi-scale feature pyramid, \cite{mao2019learning} introduced a novel attention framework for VPR to re-weight and select distinctive features, however, its performance depended on the extent of illumination variations. 

SPE-NetVLAD~\cite{8700608} enhanced NetVLAD~\cite{arandjelovic2016netvlad} by encoding the last convolutional layer using multiple patch sizes. However, the consequent concatenation of several patch-level VLAD vectors results in viewpoint-sensitivity and undesirably high-dimensional image representation. %
Being recent and high-performing global descriptor technique due to the use of VLAD aggregation, we include this method in our benchmark comparisons. Furthermore, we also show that its performance can be further enhanced when used in conjunction with our proposed method, since, instead of learning from multiple patches on the last layer, we use multiple low-resolution images to perform a multi-resolution VLAD aggregation, resulting in a superior-performing descriptor while being the same size as the original NetVLAD.

\begin{figure*}
	\centering
	\includegraphics[trim={0 1cm 0 0.25cm},clip,width=\textwidth]{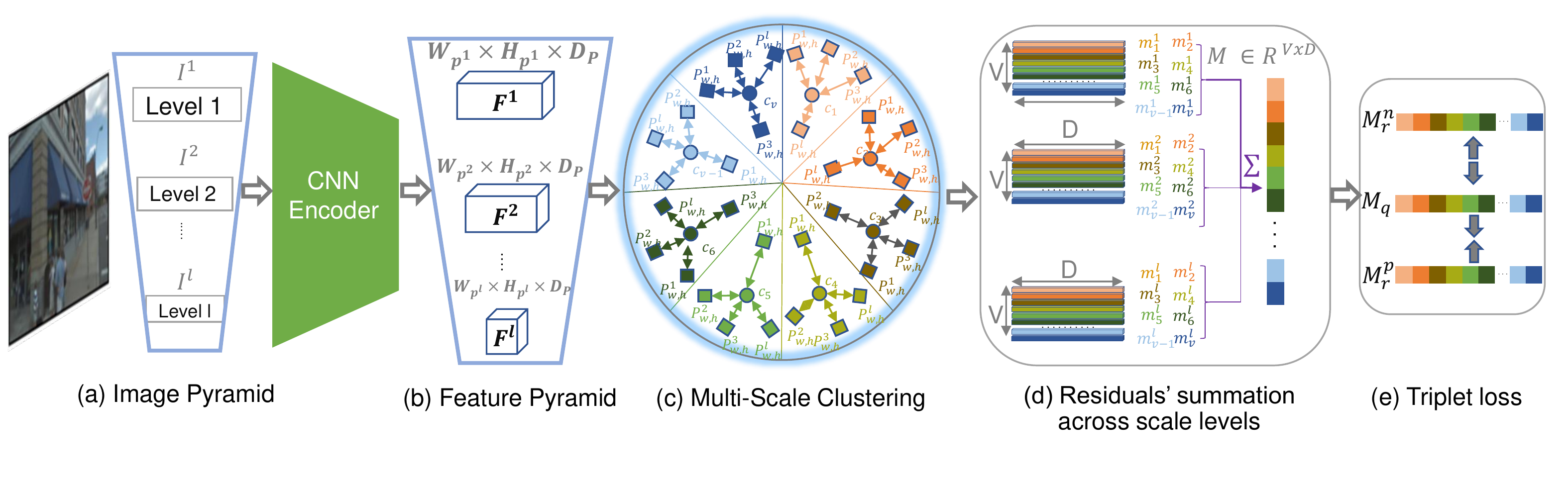}
	\caption{An overview of our proposed low-resolution image pyramid based NetVLAD training.}
	\label{figure:proposedMethod}
\end{figure*}

\section{METHODOLOGY}
\label{section:prop_method}
In this section, we describe the proposed low-resolution image pyramid based feature learning method, as illustrated in Figure~\ref{figure:proposedMethod}. We first introduce the notation for the low-resolution image/feature pyramid, then discuss the joint multi-resolution VLAD aggregation process which is trained with triplet loss, and finally present the global descriptor variants that are suited to matching across varying viewpoint conditions.

\subsection{Low-Resolution Image/Feature Pyramid}
Given a CNN as an image encoder with only the convolutional layers retained, it maps $I^l$ input image tensor to $ W_{P^l} \times H_{P^l}  \times D_{P} $, where $ W $, $ H $, and $D$ represent the width, height and number of channels (or feature maps); $I$ and $P$ denote the image and feature space; $l \in \mathbb{Z}$ is the resolution reduction factor from a pre-defined set, cardinality of which is denoted by $L$.
We consider a base resolution $BR=I^1$ ($W_I=640, \ H_I=480$ and $l=1$ in this case) and then construct a low resolution image pyramid using $l$, where $I^l$ represents $l^{th}$ resolution-reduced image, as below:

\begin{equation}
I^l = (\ \frac{W_I}{l},\  \frac{H_I} {l},\ 3 \ ) 
\label{eq:1}
\end{equation}

For the input image pyramid $I^l$, we obtain a corresponding output feature pyramid $P^l$, as shown in Figure~\ref{figure:proposedMethod}a and~\ref{figure:proposedMethod}b. While it is a common practice in the CNN literature to construct such image/feature pyramids, the above notation and details are important for the multi-resolution VLAD aggregation which has not been previously explored in the literature, which is detailed in the following section.

\subsection{Multi-Resolution Residual Aggregation}

To enable learning of a compact image representation, we introduce a trainable \textit{MultiRes-NetVLAD} layer that jointly encodes the multi-resolution $P^l$ features into $V \times D$ dimensional space, where $v \in \{1,2,3..., V\}$ represents the cluster centers as the shared feature vocabulary for all the resolutions
(see Figure~ \ref{figure:proposedMethod}c).
Similar to the original NetVLAD implementation and as shown in Figure~\ref{figure:proposedMethod}c and~\ref{figure:proposedMethod}d, for each resolution level $l$, softmax assignment $S^l_{w,h,v}$ is computed for each cluster center $c_v$ to obtain $D$-dimensional residual $m^l_v$:

\begin{equation}
m^l_v = \sum\limits_{w=1}^W \sum\limits_{h=1}^H S^l_{w,h,v} (P^l_{w,h} - c_v)
\label{eq:5}
\end{equation}
\begin{equation}
m_v = \sum\limits_{l=1}^L  m^l_v
\label{eq:6}
\end{equation}
where $m_v$ represents the aggregated residual across all $L$ resolutions for any given cluster center. To obtain the global VLAD representation $M$, all $L$ resolutions-aggregated $m_v$ residuals are first independently $L_2$-normalized along $D$-dimension (intra-normalization), and $L_2$-normalized after concatenating all the cluster residuals, following the standard practise to deal with visual word burstiness~\cite{farandjelovic2013all}. 

In summary, we propose to  pass multiple down-scaled versions of an image through the network, and construct a NetVLAD representation from the union of the $D$-dimensional encoder features obtained for multiple scales. This is the same as the original NetVLAD implementation~\cite{arandjelovic2016netvlad} but using a larger pool of features extracted additionally from lower image resolutions. We also considered \textit{scale-specific} clustering and aggregation but it led to performance degradation as compared to the original NetVLAD and our proposed method.

\subsection{Training with Triplet Loss}

To train the proposed method, we use max-margin triplet loss $t$, as proposed in ~\cite{arandjelovic2016netvlad}, where the objective is to minimize the distance between similar query-database image pairs ($M_q$ and $M_r^{p}$) while maximizing the distance between dissimilar pairs ($M_q$ and $M_r^{n}$), as shown in Figure~\ref{figure:proposedMethod}e. 

\begin{equation}
t = \max
([\phi (M_q,M^{p}_r) + g - \phi (M_q,M^{n}_r)],0)
\label{eq:8}
\end{equation}
where $\phi$ represents the Euclidean distance function and $g$ represents the margin. Training parameters details are provided in Section~\ref{sec:trainParams}.

\subsection{Test-time Global Descriptor Variants}
\label{approach:test-variants}
Although we train NetVLAD with multiple low-resolution images, the trained model itself is not strictly tied to the use of multiple resolutions at test time, since neither the underlying CNN nor the VLAD layer have any resolution-specific architectural modifications. Exposing the network to the low-resolution image pyramid during training aids learning of novel features which improve the representational capacity of the CNN, which can then be used as a backbone for different types of feature aggregation/pooling strategies. We refer to this MR-NetVLAD backbone as \textit{Base Resolution (BR)}  when testing it with only a single image resolution. On top of BR, we define two multi-scale variants for testing, which are motivated by their inherent viewpoint-robustness characteristics:

\subsubsection{BR + Multi-Low-Resolution (MLR) for Viewpoint-Varying Matching} Similar to the original NetVLAD, MR-NetVLAD's feature aggregation from the last convolutional layer is permutation-invariant, regardless of how many image resolutions are used. This grants NetVLAD the robustness to variations in camera viewpoints. The BR + MLR variant explicitly refers to the use of multiple low image resolutions at \textit{test time}, and is expected to perform well on viewpoint-varying datasets.

\subsubsection{BR + SPC (Spaial Pyramid Concatenation) for Viewpoint-Consistent Matching} 
SPC refers to the multi-scale VLAD descriptor concatenation proposed in SPE-NetVLAD framework~\cite{8700608}. Here, a single image resolution is used and several VLAD descriptors are concatenated together by considering multiple patch sizes on top of the final convolutional tensor (see ~\cite{8700608} for details). The use of large patch sizes and the consequent concatenation leads to reduced viewpoint-invariance but enables better matching of viewpoint-consistent datasets under significant appearance variations.

\section{EXPERIMENTAL SETUP}
\label{section:imp_details}

This section presents the implementation and evaluation details including benchmark datasets and existing state-of-the-art VPR methods.

\subsection{Benchmark Datasets}
\label{subsec:datasets}
To evaluate \textit{MR-NetVLAD} and existing VPR methods, we experimented on five large-scale place recognition datasets that exhibit challenging viewpoint and appearance variations: Pittsburgh-250k~\cite{arandjelovic2016netvlad}, Pittsburgh-30k~\cite{torii2013visual}, Berlin Kudamm~\cite{khaliq2018holistic}, Tokyo24/7~\cite{torii201524} and Oxford~\cite{10.1007/978-3-319-49409-8_74}. Other information including image-level distribution and environmental variations are detailed in Table~\ref{table:datasets}. Additionally, we use VPR-Bench~\cite{zaffar2020vpr} and report average performance across its datasets (except the already considered Pittsburgh and Tokyo24/7), which are classified as either viewpoint-varying (\textit{17-Places}, \textit{Corridor}, \textit{Essex3in1}, \textit{GardensPoint}, \textit{INRIA Holidays} and \textit{Living-room}) or viewpoint-consistent (\textit{SPEDTest}, \textit{Synthia-NightToFall}, \textit{Nordland} and \textit{Cross-seasons}).

\renewcommand{\tabcolsep}{2pt}
\begin{table}
\centering
	\caption{Benchmark place recognition datasets}
	\label{table:datasets}
\begin{tabular}{|c|c|c|c|c|}
\hline
\multirow{2}{*}{\textbf{Dataset}}                                 & \multirow{2}{*}{\textbf{Environment}}                                            & \multirow{2}{*}{\textbf{Set}} & \multirow{2}{*}{\textbf{\begin{tabular}[c]{@{}c@{}}Query\\ `Q'\end{tabular}}} & \multirow{2}{*}{\textbf{\begin{tabular}[c]{@{}c@{}}Database\\ `R'\end{tabular}}} \\
                                                                  &                                                                                  &                               &                                                                               &                                                                                  \\ \hline
\multirow{2}{*}{\textit{Pitts250k}}                               & \multirow{7}{*}{\begin{tabular}[c]{@{}c@{}}Viewpoint-\\ Varying\end{tabular}}    & Val                           & 7608                                                                          & 78648                                                                            \\ \cline{3-5} 
                                                                  &                                                                                  & Test                          & 8280                                                                          & 83952                                                                            \\ \cline{1-1} \cline{3-5} 
\multirow{3}{*}{\textit{Pitts30k}}                                &                                                                                  & Train                         & 7416                                                                          & 10000                                                                            \\ \cline{3-5} 
                                                                  &                                                                                  & Val                           & 7608                                                                          & 10000                                                                            \\ \cline{3-5} 
                                                                  &                                                                                  & Test                          & 6816                                                                          & 10000                                                                            \\ \cline{1-1} \cline{3-5} 
\textit{\begin{tabular}[c]{@{}c@{}}Kudamm\end{tabular}} &                                                                                  & -                             & 280                                                                           & 314                                                                              \\ \cline{1-1} \cline{3-5} 
\textit{Tokyo24/7}                                                &                                                                                  & Test                          & 315                                                                           & 75984                                                                            \\ \hline
\multirow{3}{*}{\textit{Oxford}}                              & \multirow{3}{*}{\begin{tabular}[c]{@{}c@{}}Viewpoint-\\ Consistent\end{tabular}} & Day-Overcast \& Night         & 1387                                                                          & 1589                                                                             \\ \cline{3-5} 
                                                                  &                                                                                  & Day-Snow \& Night             & 1580                                                                          & 1589                                                                             \\ \cline{3-5} 
                                                                  &                                                                                  & Day-Snow \& Day-Overcast         & 1580                                                                          & 1387                                                                             \\ \hline
\end{tabular}
\end{table}

\renewcommand{\tabcolsep}{3pt}
	\begin{table*}[ht]
\centering
\caption{Recall@N comparison of \textit{MR-NetVLAD} variants against state-of-the-art VPR techniques on challenging benchmark datasets with the best recall  \textbf{bolded} and second-best \textit{italicized}.
}
\label{table:sota_comparison}
\begin{tabular}{|l|c|c|c|c|l|l|l|c|l|l|l|c|l|l|l|c|}
\hline
\multirow{3}{*}{\textbf{\begin{tabular}[c]{@{}c@{}}Datasets/Methods\end{tabular}}} & \multicolumn{15}{c|}{\textbf{Viewpoint-Varying}} &\textbf{Viewpoint-Consistent}                                                                                           \\ \cline{2-17} 
                                                                                          & \textit{\begin{tabular}[c]{@{}c@{}}Pitts250k\\ Val\end{tabular}} & \textit{\begin{tabular}[c]{@{}c@{}}Pitts250k\\ Test\end{tabular}} & \textit{\begin{tabular}[c]{@{}c@{}}Pitts30k\\ Val\end{tabular}} & \multicolumn{4}{c|}{\textit{\begin{tabular}[c]{@{}c@{}}Pitts30k\\ Test\end{tabular}}} & \multicolumn{4}{c|}{\textit{Kudamm}}   & \multicolumn{4}{c|}{\textit{\begin{tabular}[c]{@{}c@{}}Tokyo24/7\\ Test\end{tabular}}} & \textit{Oxford (Avg.)} \\
                                                                                        \cline{2-17} 
                                                                                          & {R@1/5/20}                                                & {R@1/5/20}                                                 & {R@1/5/20}                                               & \multicolumn{4}{c|}{{R@1/5/20}}                                                & \multicolumn{4}{c|}{{R@1/5/20}} & \multicolumn{4}{c|}{{R@1/5/20}}                                                 & {R@1/5/20}                                   
                                                                                        \\ \hline
\textit{NetVLAD (NV) {\cite{arandjelovic2016netvlad}}}                                                  & 86.5/93.4/96.3           & 83.8/91.9/95.1    & 88.5/96.0/98.7            & \multicolumn{4}{c|}{85.4/92.9/96.2}    & \multicolumn{4}{c|}{40.4/\textit{60.7}/\textbf{81.8}}                   & \multicolumn{4}{c|}{\textit{67.0}/\textit{79.1}/\textit{86.7}}     & 66.9/80.3/91.2              \\ \hline
\textit{AP-GeM {\cite{revaud2019learning}}}                                                        & 78.5/90.0/94.4           & 79.9/90.8/95.3    & 80.7/92.7/97.2            & \multicolumn{4}{c|}{80.7/91.4/96.0}    & \multicolumn{4}{c|}{41.4/55.7/71.8}                   & \multicolumn{4}{c|}{58.4/69.8/79.4}     & 63.1/77.8/86.8              \\ \hline
\textit{DenseVLAD {\cite{torii201524}}}                                                    & 82.9/90.5/94.0           & 81.3/90.3/94.5          & 85.2/93.8/97.3            & \multicolumn{4}{c|}{80.0/90.2/95.1}    & \multicolumn{4}{c|}{40.4/57.5/77.5}                   & \multicolumn{4}{c|}{57.8/67.3/75.6}     & 51.7/64.4/78.4              \\ \hline
\textit{NV + SPC {\cite{8700608}}}                                                & 84.7/92.7/95.6                        & 84.8/92.5/95.7                 & 86.8/95.4/98.6            & \multicolumn{4}{c|}{85.9/93.4/96.5}    & \multicolumn{4}{c|}{11.1/25.0/51.8}                   & \multicolumn{4}{c|}{59.1/73.2/82.2}     & \textbf{81.7}/\textit{89.1}/\textit{93.9}              \\ \hline
\multicolumn{17}{|l|}{{\textit{\textbf{MR-NetVLAD (Ours):}}}}                                                                                                                                                                                                                                                                                                        \\ \hline
\quad w. \textit{Base Reso. (BR)}                                                                                                                & \textit{86.7}/\textit{93.5}/\textit{96.4}           & 84.9/91.9/95.6    & \textit{88.7}/96.2/98.9            & \multicolumn{4}{c|}{86.1/93.2/96.4}    & \multicolumn{4}{c|}{\textit{43.6}/\textbf{62.1}/76.1}                   & \multicolumn{4}{c|}{61.6/78.4/86.4}     & 67.0/79.7/90.4              \\ \hline
\quad w. \textit{\begin{tabular}[c]{@{}c@{}} BR + SPC\end{tabular}} & 85.6/93.0/95.9                        & \textit{85.6}/\textit{93.2}/\textit{95.9}                 & 87.4/\textit{96.4}/\textbf{99.0}            & \multicolumn{4}{c|}{\textit{86.4}/\textit{93.8}/\textit{96.7}}    & \multicolumn{4}{c|}{14.6/31.8/53.2}                   & \multicolumn{4}{c|}{61.3/75.2/85.4}     & \textit{81.5}/\textbf{89.2}/\textbf{94.3}             \\ \hline
\quad w. \textit{BR + Multi-Low-Reso.}                                                         & \textbf{88.0}/\textbf{94.1}/\textbf{97.0}           & \textbf{86.7}/\textbf{93.6}/\textbf{96.0}    & \textbf{89.4}/\textbf{96.6}/\textit{98.9}            & \multicolumn{4}{c|}{\textbf{86.8}/\textbf{93.8}/\textbf{96.7}}    & \multicolumn{4}{c|}{\textbf{44.6}/59.3/\textit{79.0}}                   & \multicolumn{4}{c|}{\textbf{69.8}/\textbf{81.3}/\textbf{88.0}}     & 68.6/80.5/90.3              \\ \hline
\end{tabular}
\end{table*}

\subsection{Benchmark Comparison Methods}

We report the performance of $15$ state-of-the-art VPR methods including those based on multi-scale feature learning techniques.
\subsubsection{Proposed MulitRes-NetVLAD}
We use the ImageNet-pretrained VGG-16~\cite{simonyan2014very} cropped at $5^{th}$ convolutional block as the image encoder. $4^{th}$ and $5^{th}$ convolution blocks are fine-tuned on Pits30k-train set with a base resolution of $640 \times 480$. From here, we refer to our method as \textbf{MR-NetVLAD}; we set $L=10$ for its training, where $l \in \{1,2,3,4,5,6,7,8,9,10\}$. Regardless of any multi-resolution setting, the proposed method always outputs an image representation with descriptor size $32768$ ($=64 \times 512$). We use PCA and whitening~\cite{peng2021semantic} as a standard practice [60] and for a fair comparison to existing approaches, thus reducing the descriptors to $4096$ dimensions. 

\subsubsection{Existing Techniques}
We use the following state-of-the-art methods in our benchmark comparisons.
\textbf{NetVLAD (NV)} \cite{arandjelovic2016netvlad}: VGG-16 based VLAD pooling, training the $4^{th}$ and $5^{th}$ convolution blocks on Pitts30k-train set (descriptor size=$32768$). 
\textbf{AP-GeM}~\cite{revaud2019learning}: optimized for average precision using generalized mean pooling (GeM) and a list-wise loss (descriptor size=$2048$). \textbf{DenseVLAD}~\cite{torii201524}: dense SIFT features sampled across the image at four different scales and aggregated using VLAD (descriptor size=$4096$). \textbf{NV + SPC}: multi-scale patches based VLAD concatenation mechanism as proposed in \textit{\textbf{SPE-NetVLAD}}~\cite{8700608}. Since no open-source code is available for \textit{NV + SPC}, we re-implement and train it as a feature aggregation technique on top of the final \textit{conv5\_3} layer of VGG-16, as described in the original work~\cite{8700608}. We do not employ their weighted triplet loss to enable a fair comparison of NetVLAD representation enhancements while keeping other things equal. Thus, we train both \textit{NV + SPC} and our proposed MR-NetVLAD using the same triplet loss as the original  NetVLAD. Furthermore, we use multi-scale levels: $3 \ $ (patches = $21$) for \textit{NV + SPC} which leads to descriptor size =$32768 \times 21$ respectively. However, PCA and whitening reduce down the image representation to $4096$ dimensions. 
We also compare our proposed methods against the $10$ techniques used in VPR-Bench~\cite{zaffar2020vpr} and report average recall performance on $10$ of its datasets.

\subsection{Evaluation}
We use \textit{Recall@N} as the evaluation metric, casting VPR as an image retrieval problem~\cite{garg2021your} that typically enables subsequent metric localization. This is in line with several existing works~\cite{arandjelovic2016netvlad,Garg2021SeqNetLD,Zhu_2018,jin2017learned,hausler2021patch}. For a query to be considered as correctly matched, we use a localization radius (in meters) of $50$ for Kudamm, $40$ for Oxford and $25$ for Pittsburgh and Tokyo. 
The pre-computed matched results and ground truth information for VPR-Bench datasets are directly borrowed from~\cite{zaffar2020vpr}.

\subsection{Training Parameters}
\label{sec:trainParams}
For all our proposed method variants and the re-implemented baselines, we use the same training parameters as the original work~\cite{arandjelovic2016netvlad} using a PyTorch re-implementation\footnote{https://github.com/Nanne/pytorch-NetVlad} and as listed here: margin $g=0.1$, clusters centers (vocabulary size) $V=64$, total training epochs $35$, optimized using SGD with $0.9$ momentum and $0.001$ weight decay, and $0.0001$ learning rate decayed by $0.5$ every $5$ epochs. For triplet set mining, we have used the same methodology of NetVLAD: training and selection of the query-positive-negative triplets are carried using weakly supervised GPS data; for a single query image $q$, $1$ positive (within $10$m) and $10$ negatives (far away than $25$m) are selected from a pool of randomly sampled $1000$ negatives; finally, hard negatives are tracked over epochs and used along with new hard negatives for training stability. %

\section{RESULTS AND DISCUSSION}
\label{section:results}
This section presents results for our proposed multi-resolution place representation methods, compared against state-of-the-art techniques. We then discuss the suitability of feature aggregation methods to viewpoint variations, increased test-time robustness to variations in the base resolution, the role of different multi-resolution configurations for training, suitable scale settings, and qualitative matches.

\subsection{Comparison against Benchmark Techniques}

Table~\ref{table:sota_comparison} %
presents Recall@N performance comparisons against several state-of-the-art methods on a range of datasets. 

\subsubsection{Richer Representation} Our proposed augmentation of NetVLAD training with a low-resolution image pyramid leads to a richer representation. This is evident from the results of multi-resolution trained MR-NetVLAD tested with only a single base resolution BR (third last row in Table~\ref{table:sota_comparison} and \ref{table:vpr_bench}), which outperforms single-resolution trained vanilla NetVLAD in most cases. Ceteris paribus, this particular comparison shows that training with multiple low image resolutions alone helps improve the representational capacity of the underlying CNN. Moreover, this added performance benefit comes at no additional computation cost during testing since only a single image resolution is used.

\subsubsection{Test-time Descriptor Variants \& Viewpoint Robustness}
As explained in Section~\ref{approach:test-variants}, using MR-NetVLAD as the backbone, its base resolution version can be combined with two variants of multi scale/resolution feature aggregation: BR + SPC for viewpoint-\textit{consistent} matching and BR + MLR for viewpoint-\textit{varying} matching. 

\textit{Viewpoint-Consistent Datasets:} In Table~\ref{table:sota_comparison}, it can be observed that BR + SPC (second last row) outperforms its vanilla counterpart NV + SPC (fourth row) on the Oxford dataset, where appearance variations are more challenging and the viewpoint remains mostly consistent across different traverses. This shows that the previously existing multi-scale feature aggregation techniques are complementary to our proposed multi-resolution MR-NetVLAD training, and their combination leads to state-of-the-art performance for the \textit{viewpoint-consistent} setting (last column of Table~\ref{table:sota_comparison} and \ref{table:vpr_bench}). \ifarxiv Performance evaluation on individual pairs of traverses of the Oxford dataset is reported in Section~\ref{subsection:oxford}.\fi

\textit{Viewpoint-Varying Datasets:} On Pitts250k, Pitts30k, Kudamm and Tokyo24/7, it can be observed that either BR or BR + MLR achieve state-of-the-art performance, while consistently outperforming BR + SPC. This contrast in the performance of the proposed descriptor variants is in line with their inherent design which suits either viewpoint-varying or viewpoint-consistent datasets, providing users with a choice based on the end application type. The performance variation is generally high across the experiments but there is a consistent performance gain for BR + MLR setting. For Kudamm, performance deteriorates for recall at 5/20 under BR-only and BR + MLR setting. This can be attributed to the significant lack of visual overlap between the reference and queries of this dataset captured respectively from a footpath and bus, which is quite different from a more systematic viewpoint variations captured in the Pittsburgh training dataset. For Tokyo24/7, there is a drop in recall under BR-only configuration, although BR + MLR achieves the best results. This can be attributed to simultaneous variations in viewpoint and appearance (day-night), which is particular to this dataset.

\begin{table}
\centering
\caption{Average Recall@N performance evaluation of VPR frameworks on VPR-Bench datasets~\cite{zaffar2020vpr} with the best recall \textbf{bold}+\underline{underlined}, second-best recall \textbf{bolded} and third-best \textit{italicized}. 
}
\label{table:vpr_bench}
\begin{tabular}{|ccc|}
\hline
\multicolumn{1}{|c|}{\multirow{3}{*}{\textbf{\begin{tabular}[c]{@{}c@{}}VPR-Bench \\ Datasets/\\ Techniques\end{tabular}}}} & \multicolumn{1}{c|}{\textit{\textbf{Viewpoint-Varying}}} & \textit{\textbf{Viewpoint-Consistent}} \\ \cline{2-3} 
\multicolumn{1}{|c|}{}                                                                                                      & \multicolumn{1}{c|}{\multirow{2}{*}{R@1/5/20}}           & \multirow{2}{*}{R@1/5/20}              \\
\multicolumn{1}{|c|}{}                                                                                                      & \multicolumn{1}{c|}{}                                    &                                        \\ \hline
\multicolumn{1}{|c|}{\textit{RegionVLAD}}                                                                                   & \multicolumn{1}{c|}{54.6/77.0/93.3}                      & 52.8/65.2/73.5                         \\ \hline
\multicolumn{1}{|c|}{\textit{CoHOG}}                                                                                        & \multicolumn{1}{c|}{61.9/79.0/87.7}                      & 43.1/53.3/64.9                         \\ \hline
\multicolumn{1}{|c|}{\textit{HOG}}                                                                                          & \multicolumn{1}{c|}{26.8/39.0/54.5}                      & 52.2/58.9/67.2                         \\ \hline
\multicolumn{1}{|c|}{\textit{AlexNet}}                                                                                      & \multicolumn{1}{c|}{40.2/55.2/72.6}                      & 52.0/59.7/68.2                         \\ \hline
\multicolumn{1}{|c|}{\textit{AMOSNet}}                                                                                      & \multicolumn{1}{c|}{53.6/72.2/88.0}                      & \textbf{71.8}/\underline{\textbf{80.3}}/\underline{\textbf{86.8}}                         \\ \hline
\multicolumn{1}{|c|}{\textit{HybridNet}}                                                                                    & \multicolumn{1}{c|}{56.3/74.7/87.7}                      & \textit{71.0}/77.6/\textbf{83.7}                \\ \hline
\multicolumn{1}{|c|}{\textit{CALC}}                                                                                         & \multicolumn{1}{c|}{27.5/45.0/67.7}                      & 45.2/55.5/64.3                         \\ \hline
\multicolumn{1}{|c|}{\textit{AP-GeM}}                                                                                       & \multicolumn{1}{c|}{67.0/85.2/94.5}                      & 59.5/67.3/72.1                         \\ \hline
\multicolumn{1}{|c|}{\textit{DenseVLAD}}                                                                                    & \multicolumn{1}{c|}{\underline{\textbf{72.5}}/87.3/93.8}                      & 67.7/73.7/78.5                         \\ \hline
\multicolumn{1}{|c|}{NetVLAD (NV)}                                                                                               & \multicolumn{1}{c|}{68.1/\textit{87.6}/\textit{96.3}}                      & 69.1/73.7/76.1                         \\ \hline
\multicolumn{1}{|c|}{NV + SPC}                                                                                              & \multicolumn{1}{c|}{62.6/81.3/92.5}                      & 70.9/75.9/78.4                         \\ \hline
\multicolumn{3}{|c|}{\textit{\textbf{MR-NetVLAD (Ours):}}}                                                                                                                                                                      \\ \hline
\multicolumn{1}{|c|}{\textit{BR}}                                                                                           & \multicolumn{1}{c|}{\textit{70.1}/\textbf{87.8}/\textbf{96.3}}             & 69.7/74.8/77.5                         \\ \hline
\multicolumn{1}{|c|}{\textit{BR + SPC}}                                                                                     & \multicolumn{1}{c|}{64.2/82.9/92.4}                      & \underline{\textbf{73.6}}/\textbf{78.6}/\textit{83.4}                         \\ \hline
\multicolumn{1}{|c|}{\textit{BR + MLR}}                                                                                     & \multicolumn{1}{c|}{\textbf{72.0}/\underline{\textbf{88.8}}/\underline{\textbf{96.3}}}                      & 68.8/73.7/77.1                         \\ \hline
\end{tabular}
\end{table}

\textit{VPR-Bench Datasets \cite{zaffar2020vpr}:} 
Depending on the viewpoint variations, VPR-Bench datasets are classified into either \textit{viewpoint-varying} or \textit{viewpoint-consistent} category, as earlier described in Section~\ref{subsec:datasets}. In Table~\ref{table:vpr_bench}, under viewpoint-varying environment, our BR + MLR and BR systems achieve either the \underline{\textbf{best}} or \textbf{second best} average recall values. Under viewpoint-consistent but strong appearance variations, our BR + SPC achieves the best average R@1, with R@5/20 being second/third best. 

From Table~\ref{table:sota_comparison} and ~\ref{table:vpr_bench}, it can be inferred that our methods achieve in most cases the best, and in some cases the second best, recall performance under different viewpoint/appearance variations. 
Furthermore, the average recall performance of our BR-setting is superior to the vanilla NetVLAD, highlighting the generalization of training using multi-resolution imagery, where testing is done only with a single resolution. \ifarxiv An individual dataset-level recall performance evaluation is provided in the Section~\ref{subsection:vprbench}.\fi

\subsection{Ablations, Analyses and Visualizations}
In this section, we present analyses related to the effect of using different multi-resolution configurations, different training strategies employing low-resolution imagery, and sensitivity to variations in the base resolution. Finally, we present qualitative matches for different methods. 

\subsubsection{Effect of Different Multi-Resolution Training Configurations}
Keeping base resolution $BR = 640 \times 480$, we consider four multi-resolution configurations during training: $L=1$, which defaults to vanilla NetVLAD; $L=3$, where $l \in \{1,2,4\}$; $L=6$, where $l \in \{1,2,4,6,8,10\}$; and $L=10$, where $l \in \{1,2,3,4,5,6,7,8,9,10\}$. Table~\ref{table:all_sale_confs} presents R@1 performance comparison on a subset of the benchmark datasets. It can be observed that $L=10$ configuration achieves the best recall, where both the coverage of scale space (in terms of the lowest resolution considered) as well as the density of coverage (number of image resolutions used, $L$) are the highest among all the options considered.  \ifarxiv In Section~\ref{subsection:gaussianPyr}, we \else We \fi also conducted separate studies: one extending the choices of density ($L$) and coverage (lowest resolution) of scale-space while using a different base resolution ($256 \times 256$), and the other based on a Gaussian image pyramid. These studies reinforced the results presented here, showing that using a larger coverage of scale-space by including more low-resolution images \textit{and} increasing the density of scales used within that large coverage lead to state-of-the-art performance. %

\renewcommand{\tabcolsep}{1pt}
\begin{table}
\centering
\caption{Effect of different multi-resolution configurations with the best recall \textbf{bold}+\underline{underlined}, second-best \textbf{bolded} and third-best \textit{italicized}.
}
\label{table:all_sale_confs}
\begin{tabular}{|c|c|clll|clll|clll|c|}
\hline
\multirow{2}{*}{\textbf{\begin{tabular}[c]{@{}c@{}}MR-NetVLAD\\ (setting)\end{tabular}}} & \multirow{2}{*}{\textbf{\begin{tabular}[c]{@{}c@{}}Datasets/\\ Configuration\end{tabular}}} & \multicolumn{4}{c|}{\textit{\begin{tabular}[c]{@{}c@{}}Pitts30k\\ Test\end{tabular}}} & \multicolumn{4}{c|}{Kudamm}        & \multicolumn{4}{c|}{\textit{\begin{tabular}[c]{@{}c@{}}Tokyo24/7\\ Test\end{tabular}}} & \textit{\begin{tabular}[c]{@{}c@{}}Oxford\\ SnVsOvr\end{tabular}} \\ \cline{3-15} 
                                                                                         &                                                                                             & \multicolumn{4}{c|}{R@1}                                                              & \multicolumn{4}{c|}{R@1}           & \multicolumn{4}{c|}{R@1}                                                               & R@1                                                               \\ \hline
\textit{BR / NetVLAD}                                                                    & L=1, l$\in$\{1\}                                                                                         & \multicolumn{4}{c|}{85.4}                                                             & \multicolumn{4}{c|}{40.4}          & \multicolumn{4}{c|}{\textbf{67.0}}                                                     & \textit{97.1}                                                     \\ \hline
\multirow{3}{*}{\textit{BR + MLR}}                                                       & L=3, l$\in$\{1.2.4\}                                                                          & \multicolumn{4}{c|}{\textbf{86.4}}                                                    & \multicolumn{4}{c|}{41.1}          & \multicolumn{4}{c|}{\textit{66.0}}                                                     & \textbf{97.6}                                                     \\ \cline{2-15} 
                                                                                         & L=6, l$\in$\{1,2,4,6,8,10\}                                                                   & \multicolumn{4}{c|}{\textit{85.6}}                                                    & \multicolumn{4}{c|}{\textbf{43.2}} & \multicolumn{4}{c|}{61.3}                                                              & 96.3                                                              \\ \cline{2-15} 
                                                                                         & L=10, l$\in$\{1,2,3,...,10\}                                                                   & \multicolumn{4}{c|}{\underline{\textbf{86.8}}}                                                    & \multicolumn{4}{c|}{\underline{\textbf{44.6}}} & \multicolumn{4}{c|}{\underline{\textbf{69.8}}}                                                     & \underline{\textbf{97.9}}                                                     \\ \hline
\end{tabular}
\end{table}

\subsubsection{Effect of single-scale low-resolution NetVLAD training} Since our proposed method leverages low-resolution imagery, we conduct an ablation study where we train vanilla NetVLAD (equivalent to MR-NetVLAD with single-scale) by gradually reducing the base resolution from $100\%$ to $50\%$ and $25\%$, starting from $BR=640 \times 480$. Table~\ref{table:nv_mr_vlad_comparison} shows that 50\% and 25\% $BR$ trained systems lead to significant performance degradation across all the datasets, indicating that performance advantage of MR-NetVLAD is attributed to aggregation of multiple resolutions rather than a single low resolution alone. %
\renewcommand{\tabcolsep}{0.7pt} 
\begin{table}
\centering
\caption{Recall@N performance comparison between different strategies of training NetVLAD and MR-NetVLAD, with $BR=640 \times 480$ (best recall \textbf{bold}+\underline{underlined}, second-best \textbf{bolded} and third-best \textit{italicized})
}
\label{table:nv_mr_vlad_comparison}
\begin{tabular}{|c|c|c|c|c|c|c|}
\hline
\multirow{3}{*}{\textbf{\begin{tabular}[c]{@{}c@{}}Technique\\ (Test setting)\end{tabular}}} & \multirow{3}{*}{\textbf{\begin{tabular}[c]{@{}c@{}}Scale\\ Level\end{tabular}}} & \multirow{3}{*}{\textbf{\begin{tabular}[c]{@{}c@{}}Datasets/\\ Train\\ setting\end{tabular}}} & \textit{\begin{tabular}[c]{@{}c@{}}Pitts30k\\ Test\end{tabular}} & \textit{\begin{tabular}[c]{@{}c@{}}Kudamm\end{tabular}} & \textit{\begin{tabular}[c]{@{}c@{}}Toyko24/7\\ Test\end{tabular}} & \textit{\begin{tabular}[c]{@{}c@{}}Oxford\\ SnVsOvr\end{tabular}} \\ \cline{4-7} 
                                                                                             &                                                                                 &                                                                                               & \multirow{2}{*}{R@1}                                             & \multirow{2}{*}{R@1}                                             & \multirow{2}{*}{R@1}                                              & \multirow{2}{*}{R@1}                                              \\
                                                                                             &                                                                                 &                                                                                               &                                                                  &                                                                  &                                                                   &                                                                   \\ \hline
\multirow{4}{*}{\begin{tabular}[c]{@{}c@{}}NetVLAD\\ (BR)\end{tabular}}                      & \multirow{3}{*}{L=1}                                                            & $BR$                                                                                          & \textit{85.5}                                                    & 40.4                                                             & \underline{\textbf{67.0}}                                                     & \textit{97.1}                                                     \\ \cline{3-7} 
                                                                                             &                                                                                 & 0.5$BR$                                                                                       & 84.0                                                             & 28.9                                                             & 44.1                                                              & 94.1                                                              \\ \cline{3-7} 
                                                                                             &                                                                                 & 0.25$BR$                                                                                      & 79.3                                                             & 19.3                                                             & 25.4                                                              & 86.5                                                              \\ \cline{2-7} 
                                                                                             & \begin{tabular}[c]{@{}c@{}}L=10\\ (Random\\ Reso)\end{tabular}                  & \begin{tabular}[c]{@{}c@{}}$BR$ +\\ random \\ resizes\end{tabular}                            & 81.4                                                             & \textit{40.7}                                                    & 58.7                                                              & 94.4                                                              \\ \hline
\multirow{2}{*}{\begin{tabular}[c]{@{}c@{}}MR-NetVLAD\\ (BR)\end{tabular}}                   & L=3                                                                             & \multirow{2}{*}{\begin{tabular}[c]{@{}c@{}}$BR$ +\\ MLR\end{tabular}}                         & \textbf{85.7}                                                    & \textbf{42.5}                                                    & \textbf{65.7}                                                     & \textbf{97.3}                                                     \\ \cline{2-2} \cline{4-7} 
                                                                                             & L=10                                                                            &                                                                                               & \underline{\textbf{86.1}}                                                    & \underline{\textbf{43.6}}                                                    & \textit{61.6}                                                     & \underline{\textbf{97.8}}                                                     \\ \hline
\end{tabular}
\end{table}

\begin{figure}
	\centering
	\includegraphics[width=1.0\linewidth, height=0.75\linewidth]{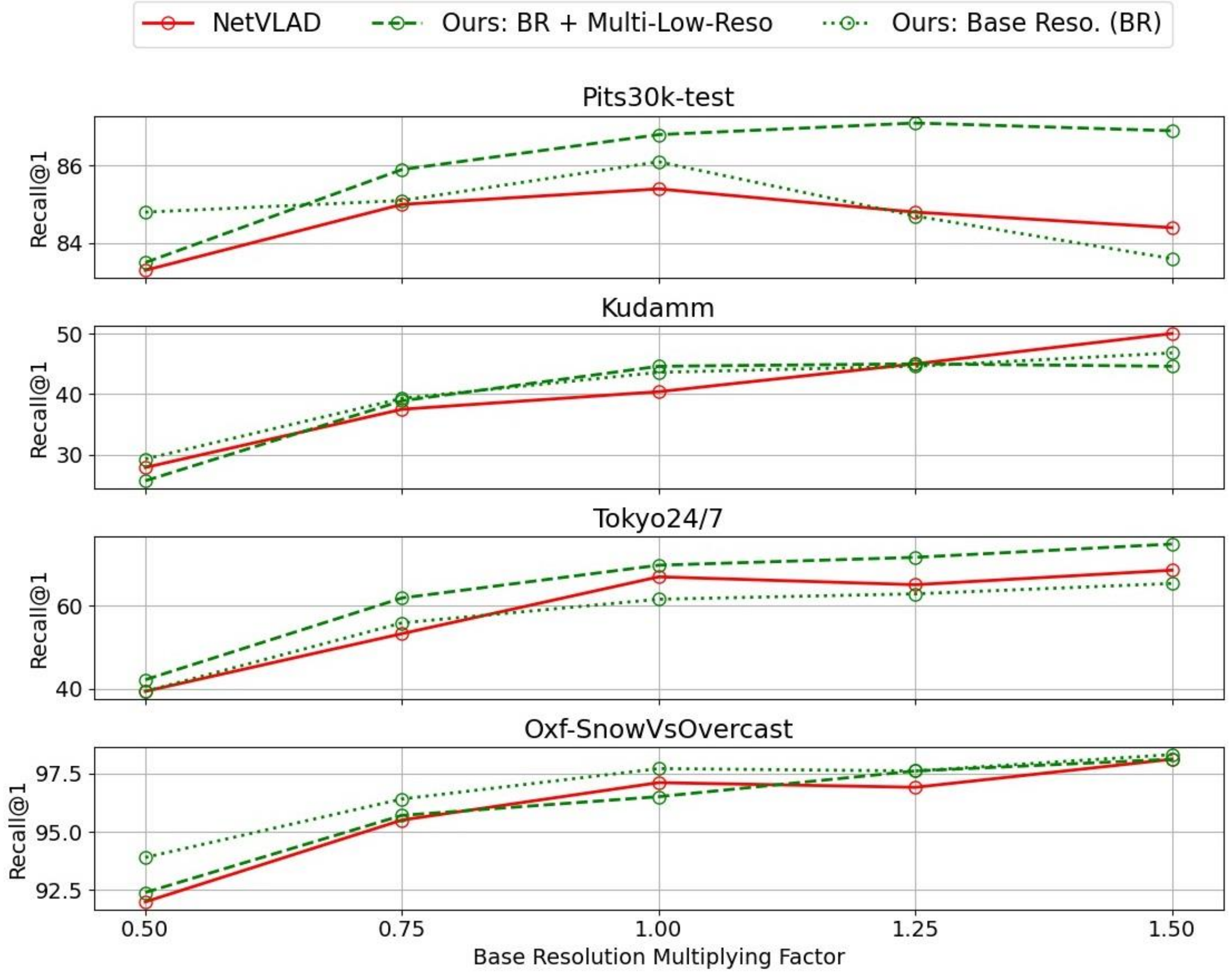}
	\caption{Sensitivity to Variations in Base Resolution.}
	\label{figure:var_baseResol}
\end{figure}

\begin{figure*}
    \centering
    \begin{tabular}{ccc}
       \includegraphics[trim={0 0cm 0 0cm},clip,width=0.3\linewidth]{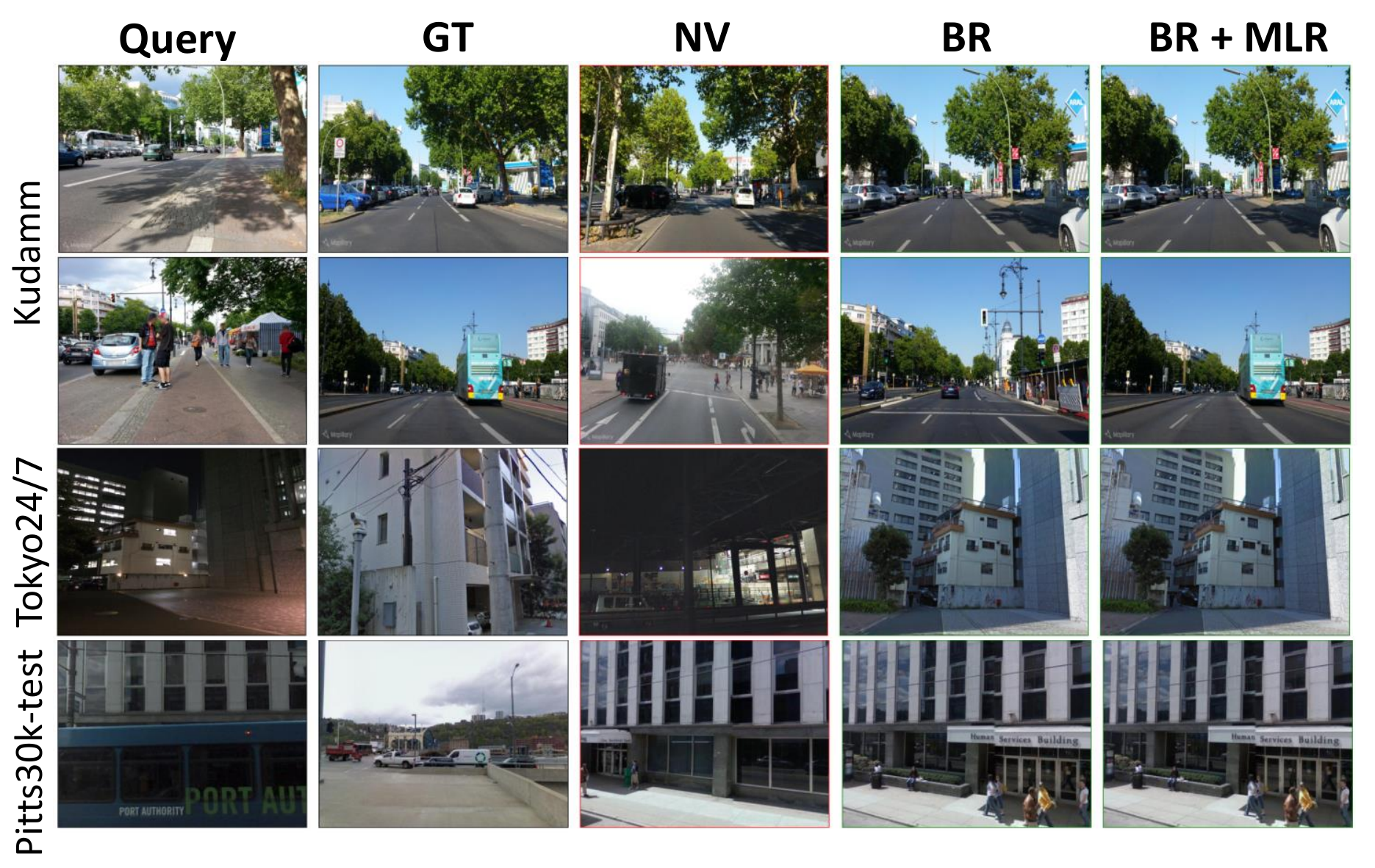} &
        \includegraphics[trim={0 0cm 0 0cm},clip,width=0.3\linewidth]{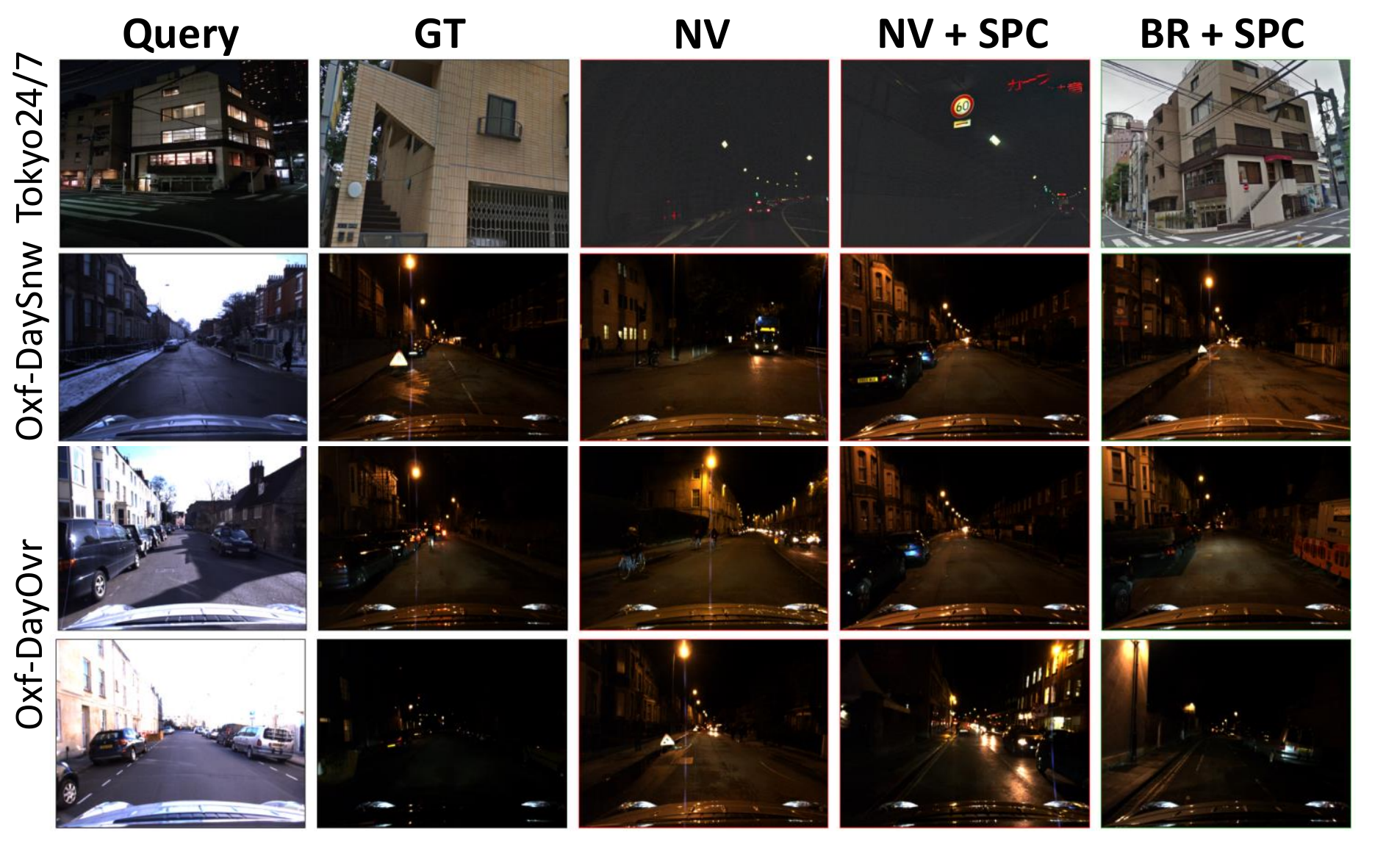} &
        \includegraphics[trim={0 0cm 0 0cm},clip,width=0.40\linewidth]{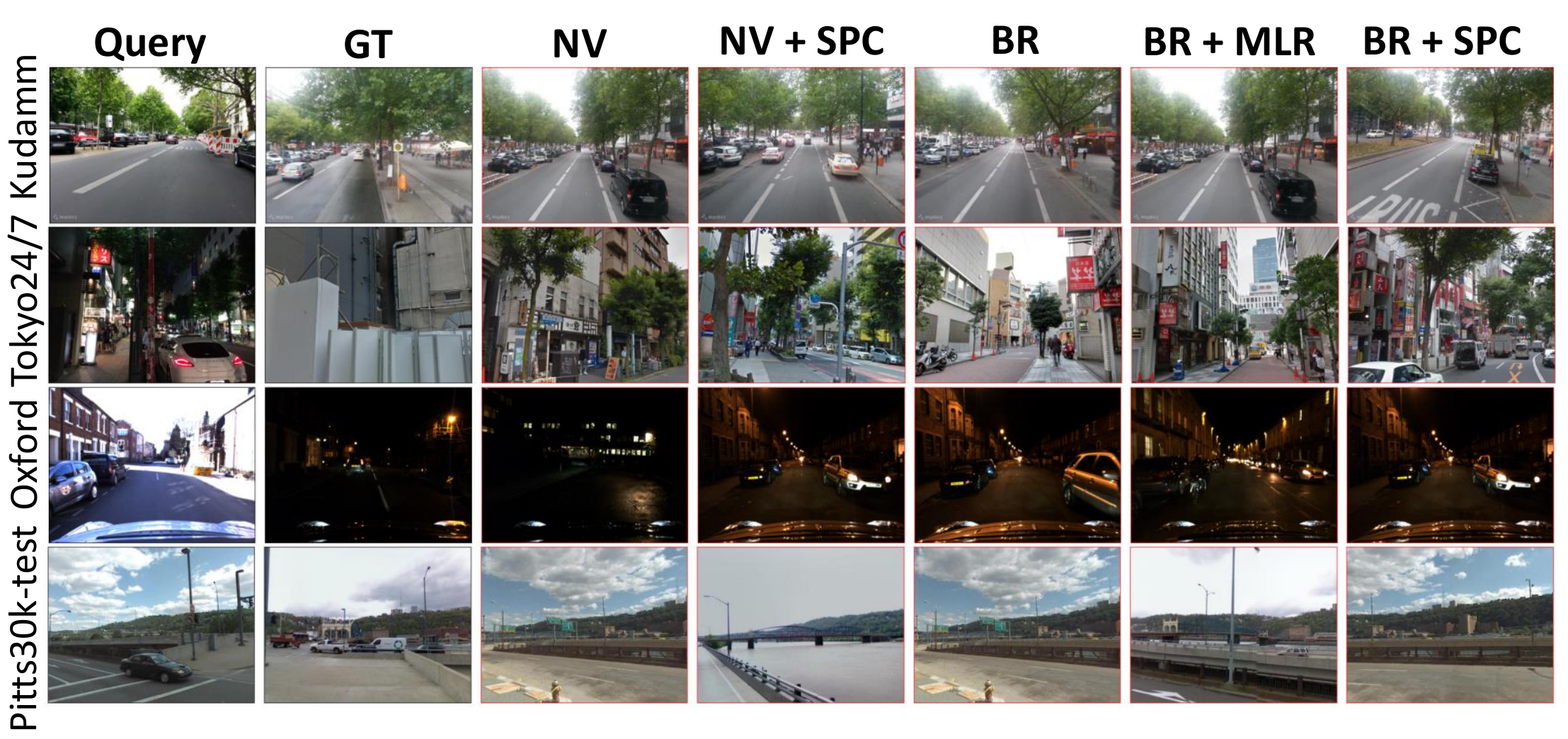} 
        \\
        (a) Correct matches & (b) Correct matches & (c) Incorrect matches \\ 
    \end{tabular}
    \caption{Qualitative results for VPR comparing our proposed method against vanilla NetVLAD, showing both correct (left, middle) and incorrect matches (right).}
    \label{figure:qual}
\end{figure*}

\subsubsection{Comparing Against Resize-Augmentation based NetVLAD training}
Here, we considered an alternative to MR-NetVLAD by utilizing multiple image resolutions through an image resizing-based data augmentation. Similar to the $L=10$ configuration, we down-sample a $BR=640 \times 480$ image with a resize factor randomly chosen from $\{1,2,3,4,5,6,7,8,9,10\}$ for every image in the batch per training iteration. In Table~\ref{table:nv_mr_vlad_comparison}, it can be observed that there is a significant performance drop (fourth row) across all the datasets except Kudamm. Since this alternative training approach forces \textit{cross-scale} VLAD comparisons of randomly-resized images, it aids in dealing with more extreme viewpoint variations as those found in Kudamm but is still worse off our proposed approach (last and second last row) which uses multi-resolution image pyramid (BR + MLR) at \textit{training time} while utilising only a single resolution (BR) at \textit{test time}. %

\subsubsection{Sensitivity to Test-Time Variations in Base Resolution}
Figure~\ref{figure:var_baseResol} shows performance comparisons on subset of benchmark datasets for variations in image resolution when testing with only a single base resolution ($BR$). The x-axis shows the factor by which the base resolution is varied, with the central value corresponding to $640 \times 480$. Note that these variations are only made during test-time and the MR-NetVLAD model used here for testing is trained on multiple image resolutions with $L=10$ configuration. Here, we consider vanilla NetVLAD and two variants of MR-NetVLAD: BR only and BR + MLR. The following observations can be made from Figure~\ref{figure:var_baseResol}: a) MR-NetVLAD's training with low-resolution image pyramid (both green lines) is less sensitive to changes in the base resolution, and consistently performs better than vanilla NetVLAD; b) for MR-NetVLAD, using multiple low resolution images \textit{at test time} (dashed green) than using only the base resolution alone (dotted green) noticeably improves performance when considering higher base resolutions, e.g. $960 \times 720$ at the rightmost end; c) BR-only setting is less sensitive to low resolutions (left side) and d) performance patterns vary across Pittsburgh vs Kudamm, which can be attributed to variations in the environment types (Pittsburgh's narrow vs Kudamm's wider roads) and camera settings such as field-of-view, both of which change the amount of visual information so captured for the same resolution setting.

\subsubsection{Qualitative Analysis}
Figure~\ref{figure:qual} shows qualitative matches including cases where our proposed methods outperform the baseline and the cases where all methods retrieved incorrect matches. These correct and incorrectly-retrieved matches are shown for the proposed MR-NetVLAD (BR, BR + MLR, BR + SPC), vanilla NetVLAD (NV) and NV + SPC ~\cite{8700608}. From Figure~\ref{figure:qual}(a) and (b), it can be seen that the proposed method is able to match places despite a forward-translational shift in the viewpoint (Kudamm, Pitts30k-test and Tokyo24/7) and strong day-night transition (Oxford sets). In the incorrect category~(c), it can be observed that failures occur due to significant perceptual aliasing caused by a similar scene structure, e.g., buildings and trees (top row), extreme appearance variations combined with structurally-similar places (second row) and challenging viewpoint change (last row), all of which still remain a challenging problem for VPR research.  

\section{CONCLUSION}
\label{section:conclus}
In this paper, we have proposed a low-resolution image pyramid based multi-resolution VLAD aggregation method, dubbed \textit{MR-NetVLAD}, for end-to-end representation learning for visual place recognition. 
Evaluation on challenging city-scale viewpoint- and appearance-varying datasets demonstrated superior recall performance of our method compared to existing state-of-the-art multi-scale place recognition methods. Results also showed that utilising multi-resolution trained \textit{MR-NetVLAD} for both single resolution testing and multi-resolution testing significantly improves performance over the vanilla NetVLAD~\cite{arandjelovic2016netvlad}. Furthermore, using the MR-NetVLAD as a backbone also improved the performance of existing multi-scale approaches such as SPE-NetVLAD~\cite{8700608} indicating the complementary nature of both the methods. Finally, we showed that MR-NetVLAD improves the robustness of global descriptors with reduced sensitivity to variations in image resolution when testing with only a single resolution. In future, we plan to extend our proposed method with local feature matching, which has been shown to benefit from multi-scale information~\cite{hausler2021patch}, but which could further leverage end-to-end learning for the VPR task.

\ifarxiv
\newpage
\section{Supplementary Material}
\label{section:suply_mat}
Here, we present additional details, evaluations, and visualizations which could not fit into the main paper due to space limitation but are valuable for any re-implementation and deeper insights.
\subsection{Multi-scale configurations for base resolution $640 \times 480$ and $256 \times 256$}
We evaluate multiple configurations of scale-space using our original \textit{n}th pixel indexing based image subsampling for base resolution $BR= 640 \times 480$ and $BR=256 \times 256$. The latter was used to keep extensive experiments computationally tractable, which also helped in tallying performance trends across base resolutions. For both the base resolutions, we gradually increase the overall coverage of the scale space by varying the lowest image resolution considered while also increasing the density ($L$) of this coverage, for example, the final $10$-scale configuration has $L=10$ images, with lowest resolution based on indexing only every $10$th pixel of the original image. 

In Table~\ref{table:all_sale_confs} and~\ref{table:low_res_test_pca}, it can be observed that $L=10$ configuration achieves the best recall in most cases, where both the coverage of scale space (in terms of the lowest resolution considered) as well as the density of coverage (number of image resolutions used, $L$) are the highest among all the options considered. In Table~\ref{table:low_res_test_pca}, it can further be observed that $l\in \{1,3,5\}$ and $l\in \{1,3,5,7,9\}$ configurations reduce recall, which can be due to a large change in scale in the first two resolutions, that is, from 1 to 1/3. In comparison, $l\in \{1,2,4\}$, $l\in \{1,2,4,6,8\}$ and $l\in \{1,2,4,6,8,10\}$ configurations still achieve comparable and better performance than the vanilla single-scale system, but their lack of density or high coverage prevents consistent top performance.

\begin{table}[b]
\centering
\caption{Effect of different multi-resolution configurations using ``$BR=640 \times 480$'' with
the best recall \textbf{bold}+\underline{underlined}, second-best \textbf{bolded} and third-best \textit{italicized}.
}
\label{table:all_sale_confs}
\begin{tabular}{|c|c|clll|clll|clll|c|}
\hline
\multirow{2}{*}{\textbf{\begin{tabular}[c]{@{}c@{}}MR-NetVLAD\\ (setting)\end{tabular}}} & \multirow{2}{*}{\textbf{\begin{tabular}[c]{@{}c@{}}Datasets / Config.\\ ($BR= 640 \times 480$)\end{tabular}}} & \multicolumn{4}{c|}{\textit{\begin{tabular}[c]{@{}c@{}}Pitts30k\\ Test\end{tabular}}} & \multicolumn{4}{c|}{Kudamm}        & \multicolumn{4}{c|}{\textit{\begin{tabular}[c]{@{}c@{}}Tokyo24/7\\ Test\end{tabular}}} & \textit{\begin{tabular}[c]{@{}c@{}}Oxford\\ SnVsOvr\end{tabular}} \\ \cline{3-15} 
                                                                                         &                                                                                                             & \multicolumn{4}{c|}{R@1}                                                              & \multicolumn{4}{c|}{R@1}           & \multicolumn{4}{c|}{R@1}                                                               & R@1                                                               \\ \hline
\textit{BR / NetVLAD}                                                                    & L=1, l$\in$ \{1\}                                                                                                          & \multicolumn{4}{c|}{85.4}                                                             & \multicolumn{4}{c|}{40.4}          & \multicolumn{4}{c|}{\textbf{67.0}}                                                     & \textit{97.1}                                                     \\ \hline
\multirow{3}{*}{\textit{BR + MLR}}                                                       & L=3, l$\in$ \{1,2,4\}                                                                                          & \multicolumn{4}{c|}{\textbf{86.4}}                                                    & \multicolumn{4}{c|}{41.1}          & \multicolumn{4}{c|}{\textit{66.0}}                                                     & \textbf{97.6}                                                     \\ \cline{2-15} 
                                                                                         & L=6, l$\in$ \{1,2,4,6,8,10\}                                                                                   & \multicolumn{4}{c|}{\textit{85.6}}                                                    & \multicolumn{4}{c|}{\textbf{43.2}} & \multicolumn{4}{c|}{61.3}                                                              & 96.3                                                              \\ \cline{2-15} 
                                                                                         & L=10, l$\in$ \{1,2,3,...,10\}                                                                                  & \multicolumn{4}{c|}{\underline{\textbf{86.8}}}                                                    & \multicolumn{4}{c|}{\underline{\textbf{44.6}}} & \multicolumn{4}{c|}{\underline{\textbf{69.8}}}                                                     & \underline{\textbf{97.9}}                                                     \\ \hline
\end{tabular}
\end{table}

\renewcommand{\tabcolsep}{1.2pt}  
\begin{table*}[ht]
\centering
\caption{Effect of different multi-resolution configurations using ``$BR=256 \times 256$'' with the best recall \textbf{bold}+\underline{underlined}, second-best \textbf{bolded} and third-best \textit{italicized}. 
}
\label{table:low_res_test_pca}
\begin{tabular}{|c|cc|c|c|c|c|}
\hline
\multirow{2}{*}{\textbf{\begin{tabular}[c]{@{}c@{}}MR-NetVLAD\\ (setting)\end{tabular}}} & \multicolumn{2}{c|}{\multirow{2}{*}{\textbf{\begin{tabular}[c]{@{}c@{}}Datasets / Configuration\\ ($BR= 256 \times 256$)\end{tabular}}}} & \textit{\begin{tabular}[c]{@{}c@{}}Pitts30k\\ Test\end{tabular}} & \begin{tabular}[c]{@{}c@{}}Kudamm \end{tabular} & \textit{\begin{tabular}[c]{@{}c@{}}Tokyo24/7\\ Test\end{tabular}} & \textit{\begin{tabular}[c]{@{}c@{}}Oxford\\ SnowVsOver\end{tabular}} \\ \cline{4-7} 
                                                                                     & \multicolumn{2}{c|}{}                                                                                                            & R@1                                                              & R@1                                                     & R@1                                                               & R@1/5/20                                                             \\ \hline
\textit{BR /  NetVLAD}                                                                   & \multicolumn{1}{c|}{L=1}                                                  & l$\in$ \{1\}                                            & \textit{82.5}                                                    & 29.6                                                    & 40.6                                                              & \textit{94.7}                                                        \\ \hline
\multirow{7}{*}{\textit{BR + MLR}}                                                       & \multicolumn{1}{c|}{\multirow{2}{*}{L=3}}                                 & l$\in$ \{1,2,4\}                                        & 82.2                                                             & \underline{\textbf{29.6}}                                           & \textit{43.5}                                                     & \textbf{95.2}                                                        \\ \cline{3-7} 
                                                                                     & \multicolumn{1}{c|}{}                                                     & l$\in$ \{1,3,5\}                                        & 81.7                                                             & 27.5                                                    & 41.9                                                              & 93.4                                                                 \\ \cline{2-7} 
                                                                                     & \multicolumn{1}{c|}{\multirow{3}{*}{L=5}}                                 & l$\in$ \{1,2,4,8,10\}                                   & \underline{\textbf{83.0}}                                                    & 26.1                                                    & 41.6                                                              & 93.2                                                                 \\ \cline{3-7} 
                                                                                     & \multicolumn{1}{c|}{}                                                     & l$\in$ \{1,3,5,7,9\}                                    & 81.6                                                             & \textit{27.9}                                           & 42.5                                                              & 93.6                                                                 \\ \cline{3-7} 
                                                                                     & \multicolumn{1}{c|}{}                                                     & l$\in$ \{1,2,3,4,5\}                                    & 81.8                                                             & \textbf{28.9}                                           & \textbf{44.8}                                                     & 93.9                                                                 \\ \cline{2-7} 
                                                                                     & \multicolumn{1}{c|}{L=6}                                                  & l$\in$ \{1,2,4,6,8,10\}                                 & 82.2                                                             & 26.4                                                    & 41.9                                                              & 94.3                                                                 \\ \cline{2-7} 
                                                                                     & \multicolumn{1}{c|}{L=10}                                                 & l$\in$ \{1,2,3,...,10\}                                 & \textbf{82.9}                                                    & 25.7                                                    & \underline{\textbf{46.7}}                                                     & \underline{\textbf{95.3}}                                                        \\ \hline
\end{tabular}
\end{table*}

\subsection{Gaussian Pyramid based multi-scale configurations for base resolution $640 \times 480$}
\label{subsection:gaussianPyr}
In this analysis, we incorporate the theoretical findings related to scale-space theory from the computer vision literature~\cite{lindeberg1994scale,burt1987laplacian,crowley2002fast,lowe2004distinctive}. Here, we present a Gaussian pyramid based image downsampling regime for analysing performance variations with respect to an expanding size of the pyramid. For this purpose, we used a fixed standard deviation ($\sigma$ ) value of 1 for a 2-D Gaussian filter, which can be approximated with sufficient accuracy using a filter size of $5 \times 5$ (encompassing approx 99\% of distribution weight in that pixel neighborhood)~\cite{lowe2004distinctive}. We construct the Gaussian pyramid by starting from the base resolution ($640 \times 480$) and then include another image after operating the 2D Gaussian filter ($\sigma=1$) and pixel subsampling by a factor $F$ using bilinear interpolation. We considered two different values of $F$: $2$ and $\sqrt{2}$ and considered the lowest image resolution to be $1/8^{th}$ of the base resolution (resulting in $5\times4$ dimensional final convolutional tensor of the VGG backbone). Thus, the two values of F present different densities of a fixed but large scale-space coverage while also varying the effective standard deviation ($\sigma_{eff}$), as follows:  

\begin{equation}
\sigma_{eff_{\textbf{i}}} = F \sqrt{\sigma_{eff_{\textbf{i-1}}}^2 +\sigma^2}
\label{eq:1}
\end{equation}

$\sigma_{eff}$ is not applicable to the base level of the pyramid and can be computed for subsequent levels starting from $\sigma_{{eff}_0}=0$.
Table~\ref{table:scale_space_theory} and Figure~\ref{figure:recalls_all} present the new results for the Gaussian pyramid based analysis. The first three columns of Table~\ref{table:scale_space_theory} respectively list the pyramid scaling factor (F), effective standard deviation ($\sigma_{eff}$) and the set of image scales in the pyramid relative to the original resolution (obtained after Gaussian filtering and subsampling), where a smaller pyramid scaling factor $\sqrt{2}$ results in a denser pyramid. Figure~\ref{figure:recalls_all} plots Recall@1/20 with respect to $\sigma_{eff}$. Through both Table~\ref{table:scale_space_theory} and Figure~\ref{figure:recalls_all}, it can be observed that: i) high performance is achieved when a large and dense pyramid is considered, that is, including more downsampled images (as we move from left to right in the Figure), and that too with high density, that is, pyramid scaling factor $\sqrt{2}$ (red) as compared to 2 (cyan); ii) performance typically deteriorates after reaching a peak value which seems to be dataset-specific and can be attributed to a high value of $\sigma_{eff}$, which is not the case with our original multi-scale configuration as no Gaussian blurring is performed; iii) performance deteriorates for the Oxford dataset, which could again be attributed to the blurring step before image subsampling; and iv) peak performance achieved through this analysis was close (in some cases better and in others worse) to what was achieved in other multi-scale analyses.

\renewcommand{\tabcolsep}{5pt} 
\begin{table*}
\centering
\caption{Recall@N for MR-NetVLAD using Gaussian Pyramid based multi-scale configurations with the best recall \textbf{bold}+\underline{underlined}, second-best recall \textbf{bolded} and third-best \textit{italicized}. 
}
\label{table:scale_space_theory}
\begin{tabular}{|ccc|c|c|c|c|}
\hline
\multicolumn{3}{|c|}{\multirow{2}{*}{\textbf{\begin{tabular}[c]{@{}c@{}}Datasets/\\ MR-NetVLAD\end{tabular}}}}                                          & \multirow{2}{*}{\textit{\begin{tabular}[c]{@{}c@{}}Pitts30k\\ Test\end{tabular}}} & \multirow{2}{*}{\textit{\begin{tabular}[c]{@{}c@{}}Kudamm\end{tabular}}} & \multirow{2}{*}{\textit{\begin{tabular}[c]{@{}c@{}}Toyko24/7\\ Test\end{tabular}}} & \multirow{2}{*}{\textit{\begin{tabular}[c]{@{}c@{}}Oxford\\ SnwVsOvr\end{tabular}}} \\
\multicolumn{3}{|c|}{}                                                                                                                                  &                                                                                   &                                                                                   &                                                                                    &                                                                                     \\ \hline
\multicolumn{1}{|c|}{Factor}                      & \multicolumn{1}{c|}{$\sigma_{eff}$} & \begin{tabular}[c]{@{}c@{}}Scale-Space\\ Pyramid\end{tabular} & R@1/5/20                                                                          & R@1/5/20                                                                          & R@1/5/20                                                                           & R@1/5/20                                                                            \\ \hline
\multicolumn{1}{|c|}{\multirow{4}{*}{2}}          & \multicolumn{1}{c|}{-}              & \{1\}                                                         & 85.4/92.9/96.2                                                                    & 40.4/60.7/81.8                                                                    & 67.0/79.1/86.7                                                                     & \underline{\textbf{97.1}}/\underline{\textbf{98.6}}/\underline{\textbf{99.7}}                                                                      \\ \cline{2-7} 
\multicolumn{1}{|c|}{}                            & \multicolumn{1}{c|}{2}              & \{1,2\}                                                       & 86.0/93.5/96.7                                                                    & 40.4/62.9/\textit{83.9}                                                                    & 64.1/78.4/87.9                                                                     & 91.3/96.4/98.2                                                                      \\ \cline{2-7} 
\multicolumn{1}{|c|}{}                            & \multicolumn{1}{c|}{2$\sqrt{2}$}    & \{1,2,4\}                                                     & 85.2/92.8/96.2                                                                    & \underline{\textbf{45.0}}/62.5/80.7                                                                    & 65.1/79.7/\underline{\textbf{92.4}}                                                                     & 92.4/\textit{97.0}/\textit{98.8}                                                                      \\ \cline{2-7} 
\multicolumn{1}{|c|}{}                            & \multicolumn{1}{c|}{2$\sqrt{21}$}   & \{1,2,4,8\}                                                   & 85.6/93.1/96.3                                                                    & 41.1/60.7/82.1                                                                    & 66.7/77.8/86.7                                                                     & 93.2/97.0/98.6                                                                      \\ \hline
\multicolumn{1}{|c|}{\multirow{7}{*}{$\sqrt{2}$}} & \multicolumn{1}{c|}{-}              & \{1\}                                                         & 85.4/92.9/96.2                                                                    & 40.4/60.7/81.8                                                                    & 67.0/79.1/86.7                                                                     & 97.1/98.6/99.7                                                                      \\ \cline{2-7} 
\multicolumn{1}{|c|}{}                            & \multicolumn{1}{c|}{$\sqrt{2}$}     & \{1,$\sqrt{2}$\}                                                       & 86.6/93.6/96.8                                                                    & 42.1/\textbf{65.0}/\underline{\textbf{84.6}}                                                                    & 67.9/83.2/89.8                                                                     & 92.2/96.1/98.2                                                                      \\ \cline{2-7} 
\multicolumn{1}{|c|}{}                            & \multicolumn{1}{c|}{$\sqrt{6}$}     & \{1,$\sqrt{2}$,2\}                                                     & 86.4/93.2/96.5                                                                    & \textbf{44.3}/\textit{62.5}/80.7                                                                    & \textit{69.5}/\textbf{84.4}/90.2                                                                     & 92.4/95.7/97.3                                                                      \\ \cline{2-7} 
\multicolumn{1}{|c|}{}                            & \multicolumn{1}{c|}{$\sqrt{14}$}    & \{1,$\sqrt{2}$,2,2$\sqrt{2}$\}                                                   & \textit{86.7}/93.6/96.5                                                                    & 39.3/57.1/80.7                                                                    & 67.9/81.9/\textit{91.4}                                                                     & 92.6/96.0/98.1                                                                      \\ \cline{2-7} 
\multicolumn{1}{|c|}{}                            & \multicolumn{1}{c|}{$\sqrt{30}$}    & \{1,$\sqrt{2}$,2,2$\sqrt{2}$,4\}                                                 & \textbf{86.8}/\textit{93.7}/\textit{96.8}                                                                    & \textit{42.9}/\underline{\textbf{66.4}}/83.2                                                                    & \textbf{69.6}/83.2/91.4                                                                     & 92.5/96.8/98.3                                                                      \\ \cline{2-7} 
\multicolumn{1}{|c|}{}                            & \multicolumn{1}{c|}{$\sqrt{62}$}    & \{1,$\sqrt{2}$,2,2$\sqrt{2}$,4,4$\sqrt{2}$\}                                               & \underline{\textbf{86.8}}/\underline{\textbf{94.2}}/\underline{\textbf{97.1}}                                                                    & 41.1/60.7/79.3                                                                    & 67.6/\textit{83.5}/89.2                                                                     & \textbf{93.4}/96.0/97.8                                                                      \\ \cline{2-7} 
\multicolumn{1}{|c|}{}                            & \multicolumn{1}{c|}{$\sqrt{126}$}   & \{1,$\sqrt{2}$,2,2$\sqrt{2}$,4,4$\sqrt{2}$,8\}                                             & 86.6/\textbf{93.7}/\textbf{97.0}                                                                    & 42.5/61.1/\textbf{84.3}                                                                    & \underline{\textbf{71.4}}/\underline{\textbf{85.7}}/\textbf{91.4}                                                                     & \textit{93.2}/\textbf{97.5}/\textbf{98.9}                                                                      \\ \hline
\end{tabular}
\end{table*}    

\begin{figure}
\centering
\includegraphics[width=\linewidth]{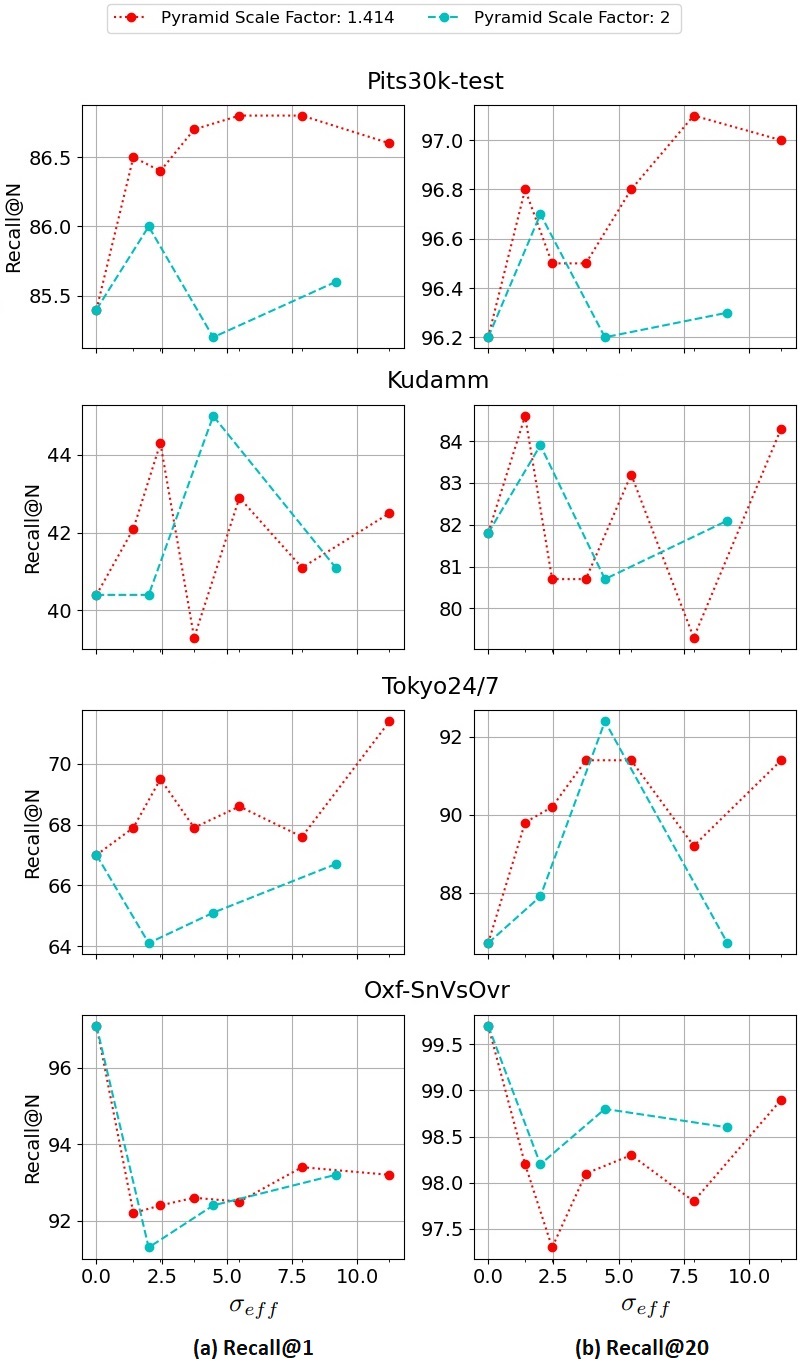}
\caption{Recall@N performance curves of the proposed \textit{MR-NetVLAD} using Gaussian Pyramid based multi-scale configurations. Although specific performance trends differ across datasets, it can be observed that high performance is achieved when a large and dense pyramid is considered, that is, including more downsampled images (as we move from left to right in any plot), and that too with high density, that is, pyramid scaling factor $\sqrt{2}$ (red) as compared to 2 (cyan). Interestingly, performance deteriorates for the Oxford dataset which can be attributed to the blurring operation since our original multi-scale configuration based on pixel subsampling alone did not have such an effect.}
\label{figure:recalls_all}
\end{figure}

\subsection{Detailed performance analysis of the Oxford dataset.}
\label{subsection:oxford}
Table~\ref{table:oxf} evaluates and presents the recall performance of dominating VPR techniques on three individual pairs of traverses of the Oxford dataset. It can be observed that our \textit{BR + SPC} and  \textit{NV + SPC} mostly achieve the \underline{\textbf{best}} and \textbf{second best} performance at both the dataset level and at an aggregate level (last column).

\begin{table*}
\centering
\caption{Recall@N performance comparison on the individual traverse pairs of the Oxford dataset with the best recall \textbf{bold}+\underline{underlined}, second-best \textbf{bolded} and third-best \textit{italicized}.
}
\label{table:oxf}    
\begin{tabular}{|c|cccc|}
\hline
\multirow{3}{*}{\textbf{\begin{tabular}[c]{@{}c@{}}Datasets/\\ Techniques\end{tabular}}} & \multicolumn{4}{c|}{\textit{Oxford-sets}}                                                                                                                 \\ \cline{2-5} 
                                                                                         & \multicolumn{1}{c|}{Day-Overcast \& Night} & \multicolumn{1}{c|}{Day-Snow \& Night} & \multicolumn{1}{c|}{Day-Snow \& Day-Overcast} & Avg            \\ \cline{2-5} 
                                                                                         & \multicolumn{1}{c|}{R@1/5/20}              & \multicolumn{1}{c|}{R@1/5/20}          & \multicolumn{1}{c|}{R@1/5/20}                 & R@1/5/20       \\ \hline
\textit{NetVLAD (NV)}                                                                         & \multicolumn{1}{c|}{\textit{53.1}/\textit{74.0}/\textit{89.4}}        & \multicolumn{1}{c|}{\textit{50.4}/\textit{68.2}/\textit{84.4}}    & \multicolumn{1}{c|}{\textit{97.1}/98.6/99.7}           & \textit{66.9}/\textit{80.3}/\textit{91.2} \\ \hline
\textit{AP-GeM}                                                                          & \multicolumn{1}{c|}{47.2/67.9/81.0}        & \multicolumn{1}{c|}{45.6/66.3/79.5}    & \multicolumn{1}{c|}{96.6/\textbf{99.1}/\underline{\textbf{99.9}}}           & 63.1/77.8/86.8 \\ \hline
\textit{DenseVLAD}                                                                       & \multicolumn{1}{c|}{37.2/54.2/73.3}        & \multicolumn{1}{c|}{27.6/43.9/64.6}    & \multicolumn{1}{c|}{90.4/95.0/97.3}           & 51.7/64.4/78.4 \\ \hline
\textit{NV + SPC}                                                              & \multicolumn{1}{c|}{\textbf{72.9}/\textbf{86.3}/\textbf{93.5}}        & \multicolumn{1}{c|}{\underline{\textbf{68.5}}/\underline{\textbf{80.2}}/\underline{\textbf{88.5}}}    & \multicolumn{1}{c|}{\underline{\textbf{98.0}}/\textit{99.0}/\textit{99.6}}           & \underline{\textbf{81.7}}/\textbf{89.1}/\textbf{93.9} \\ \hline
\textit{Ours: BR + SPC}                                                              & \multicolumn{1}{c|}{\underline{\textbf{77.8}}/\underline{\textbf{88.4}}/\underline{\textbf{95.3}}}      & \multicolumn{1}{c|}{\textbf{68.2}/\textbf{79.8}/\textbf{87.8}}    & \multicolumn{1}{c|}{\textbf{98.5}/\underline{\textbf{99.4}}/\textbf{99.7}}           & \textbf{81.5}/\underline{\textbf{89.2}}/\underline{\textbf{94.3}} \\ \hline
\end{tabular}
\end{table*}

\subsection{Performance analysis on individual VPR-Bench datasets}
\label{subsection:vprbench}
VPR-Bench~\cite{zaffar2020vpr} categorizes the datasets based on the type of surrounding environment that can either be outdoor, indoor, or a mixture of both. We explicitly consider the viewpoint variation or consistency that the particular dataset exhibits. Thus, each VPR-Bench dataset is classified into either \textit{viewpoint-varying} or \textit{viewpoint-consistent} category.

Table~\ref{table:vpr_bench} evaluates the R@1/5 performance of several techniques on VPR-Bench datasets. It can be seen that our BR and BR + Multi-Low-Reso (MLR) systems consistently achieve better and comparable recall performance across all the datasets. The only exception to this trend is the \textit{Corridor} dataset, which was rather captured under only a mild lateral viewpoint-shift, thus still preserving the overall scene structure across query/reference traverse. Thus, the patch-based approach (Spat-Pyr-Concat (SPC) / our BR + SPC) improves the recall performance significantly on \textit{Corridor}. Under challenging viewpoint-variant datasets (17-Places, GardensPoint and Living-room), our BR and BR + MLR significantly improve (\underline{\textbf{best}} and \textbf{second-best}) recall performances. For Corridor, Essex3in1 and Inria holidays sets, techniques including HybridNet, DenseVLAD and AP-GeM lead with \underline{\textbf{best}} recall performance, followed up with our proposed framework-variants that give \textit{third-best} recall performance. Furthermore, under similar viewpoint and strong appearance variations, our BR + SPC achieves superior recall performance over majority of the datasets, except the extremely challenging summer-winter transition in the \textit{Nordland} dataset where SPED-centric AMOSNet and HybridNet achieve \underline{\textbf{best}} and \textbf{second-best} performance and our BR + SPC setting boost up the recall performance by $2x$ times, achieving \textit{third-best} recall value.

\renewcommand{\tabcolsep}{3.5pt} 
\begin{table*}
\centering
\caption{Recall@N performance of VPR techniques on \textit{viewpoint-varying} and \textit{viewpoint-consistent} VPR-Bench datasets~\cite{zaffar2020vpr},  with the best recall \textbf{bold}+\underline{underlined}, second-best \textbf{bolded} and third-best \textit{italicized}.}
\label{table:vpr_bench}
\begin{tabular}{|ccccccccccc|}
\hline
\multicolumn{1}{|c|}{\multirow{3}{*}{\textit{\textbf{\begin{tabular}[c]{@{}c@{}}VPR-Bench \\ Datasets/\\ Techniques\end{tabular}}}}} & \multicolumn{6}{c|}{\textbf{Viewpoint-Varying}}                                                                                                                                                                                                                                 & \multicolumn{4}{c|}{\textbf{Viewpoint-Consistent}}                                                                                                                                                   \\ \cline{2-11} 
\multicolumn{1}{|c|}{}                                                                                                               & \multicolumn{1}{c|}{\textit{17-Places}}          & \multicolumn{1}{c|}{\textit{Corridor}}               & \multicolumn{1}{c|}{\textit{Essex3in1}}               & \multicolumn{1}{c|}{\textit{GardensPoint}}            & \multicolumn{1}{c|}{\textit{Inria Holidays}}          & \multicolumn{1}{c|}{\textit{Living-room}}           & \multicolumn{1}{c|}{\textit{Nordland}}                & \multicolumn{1}{c|}{\textit{SPEDTEST}}           & \multicolumn{1}{c|}{\begin{tabular}[c]{@{}c@{}}\textit{Synthia-}\\ \textit{NightToFall}\end{tabular}} & \textit{Cross-Season}          \\ \cline{2-11} 
\multicolumn{1}{|c|}{}                                                                                                               & \multicolumn{1}{c|}{R@1/5} & \multicolumn{1}{c|}{R@1/5}     & \multicolumn{1}{c|}{R@1/5}      & \multicolumn{1}{c|}{R@1/5}      & \multicolumn{1}{c|}{R@1/5}      & \multicolumn{1}{c|}{R@1/5}    & \multicolumn{1}{c|}{R@1/5}      & \multicolumn{1}{c|}{R@1/5} & \multicolumn{1}{c|}{R@1/5}                                             & R@1/5    \\ \hline
\multicolumn{1}{|c|}{\textit{RegionVLAD}}                                                                                            & \multicolumn{1}{c|}{39.9/61.1}     & \multicolumn{1}{c|}{43.2/71.2}         & \multicolumn{1}{c|}{59.0/83.3}          & \multicolumn{1}{c|}{43.0/74.0}          & \multicolumn{1}{c|}{80.0/88.3}          & \multicolumn{1}{c|}{62.5/84.4}         & \multicolumn{1}{c|}{6.2/13.5}           & \multicolumn{1}{c|}{56.7/69.7}     & \multicolumn{1}{c|}{62.0/81.1}                                                 & 86.4/96.3        \\ \hline
\multicolumn{1}{|c|}{\textit{CoHOG}}                                                                                                 & \multicolumn{1}{c|}{38.9/57.6}     & \multicolumn{1}{c|}{62.2/89.2}         & \multicolumn{1}{c|}{\textbf{82.4}/\textbf{92.9}}          & \multicolumn{1}{c|}{39.0/59.5}          & \multicolumn{1}{c|}{65.0/75.0}          & \multicolumn{1}{c|}{84.4/100}          & \multicolumn{1}{c|}{2.8/5.2}            & \multicolumn{1}{c|}{49.4/61.3}     & \multicolumn{1}{c|}{77.2/84.5}                                                 & 42.9/62.3        \\ \hline
\multicolumn{1}{|c|}{\textit{HOG}}                                                                                                   & \multicolumn{1}{c|}{22.4/39.4}     & \multicolumn{1}{c|}{47.7/72.1}         & \multicolumn{1}{c|}{3.8/9.5}            & \multicolumn{1}{c|}{19.0/32.0}          & \multicolumn{1}{c|}{15.3/21.3}          & \multicolumn{1}{c|}{53.1/59.4}        & \multicolumn{1}{c|}{3.4/7.7}            & \multicolumn{1}{c|}{49.9/60.1}     & \multicolumn{1}{c|}{97.7/98.0}                                                 & 57.6/69.6        \\ \hline
\multicolumn{1}{|c|}{\textit{AlexNet}}                                                                                               & \multicolumn{1}{c|}{30.0/50.7}     & \multicolumn{1}{c|}{68.5/90.1}         & \multicolumn{1}{c|}{14.3/25.7}          & \multicolumn{1}{c|}{25.0/45.0}          & \multicolumn{1}{c|}{44.0/57.0}          & \multicolumn{1}{c|}{59.4/62.5}        & \multicolumn{1}{c|}{9.2/14.7}           & \multicolumn{1}{c|}{51.6/61.6}     & \multicolumn{1}{c|}{62.0/73.6}                                                 & 85.3/89.0        \\ \hline
\multicolumn{1}{|c|}{\textit{AMOSNet}}                                                                                               & \multicolumn{1}{c|}{39.2/55.9}     & \multicolumn{1}{c|}{\textbf{84.7}/\textbf{99.1}}          & \multicolumn{1}{c|}{26.2/45.2}          & \multicolumn{1}{c|}{47.5/81.0}          & \multicolumn{1}{c|}{68.0/80.3}          & \multicolumn{1}{c|}{56.2/71.9}         & \multicolumn{1}{c|}{\underline{\textbf{29.8}}/\underline{\textbf{44.1}}}          & \multicolumn{1}{c|}{79.4/90.0}     & \multicolumn{1}{c|}{84.3/89.5}                                                 & 93.7/97.4        \\ \hline
\multicolumn{1}{|c|}{\textit{HybridNet}}                                                                                             & \multicolumn{1}{c|}{40.1/58.6}     & \multicolumn{1}{c|}{\underline{\textbf{90.1}}/\underline{\textbf{99.1}}}          & \multicolumn{1}{c|}{28.6/46.7}          & \multicolumn{1}{c|}{45.0/79.0}          & \multicolumn{1}{c|}{72.0/83.3}          & \multicolumn{1}{c|}{62.5/81.2}        & \multicolumn{1}{c|}{\textbf{19.5}/\textbf{29.6}}          & \multicolumn{1}{c|}{\textit{79.4}/\textit{90.1}}     & \multicolumn{1}{c|}{88.9/92.9}                                                 & 96.3/97.9        \\ \hline
\multicolumn{1}{|c|}{\textit{CALC}}                                                                                                  & \multicolumn{1}{c|}{30.3/46.3}     & \multicolumn{1}{c|}{32.4/48.6}         & \multicolumn{1}{c|}{11.4/22.4}          & \multicolumn{1}{c|}{17.5/40.5}          & \multicolumn{1}{c|}{33.0/46.7}          & \multicolumn{1}{c|}{40.6/65.6}        & \multicolumn{1}{c|}{3.7/7.2}            & \multicolumn{1}{c|}{42.7/56.2}     & \multicolumn{1}{c|}{68.4/84.6}                                                 & 66.0/73.8        \\ \hline
\multicolumn{1}{|c|}{\textit{AP-GeM}}                                                                                                & \multicolumn{1}{c|}{42.1/64.3}     & \multicolumn{1}{c|}{53.2/81.1}         & \multicolumn{1}{c|}{69.5/88.1}          & \multicolumn{1}{c|}{54.5/81.0}          & \multicolumn{1}{c|}{\underline{\textbf{92.3}}/\underline{\textbf{96.7}}}          & \multicolumn{1}{c|}{90.6/100}          & \multicolumn{1}{c|}{4.9/7.8}            & \multicolumn{1}{c|}{51.7/69.5}     & \multicolumn{1}{c|}{86.5/92.7}                                                 & 94.8/99.0         \\ \hline
\multicolumn{1}{|c|}{\textit{DenseVLAD}}                                                                                             & \multicolumn{1}{c|}{43.8/62.8}     & \multicolumn{1}{c|}{70.3/95.5}          & \multicolumn{1}{c|}{\underline{\textbf{91.0}}/\underline{\textbf{99.0}}}          & \multicolumn{1}{c|}{47.5/68.5}          & \multicolumn{1}{c|}{\textbf{88.3}/92.0}          & \multicolumn{1}{c|}{93.8/\textit{100}}          & \multicolumn{1}{c|}{7.4/13.7}           & \multicolumn{1}{c|}{72.8/86.5}     & \multicolumn{1}{c|}{91.1/95.0}                                                 & 99.5/99.5         \\ \hline
\multicolumn{1}{|c|}{\textit{NetVLAD (NV)}}                                                                                               & \multicolumn{1}{c|}{\textit{44.3}/\textbf{67.0}}     & \multicolumn{1}{c|}{60.4/86.5}         & \multicolumn{1}{c|}{68.6/89.5}          & \multicolumn{1}{c|}{61.5/90.0}          & \multicolumn{1}{c|}{86.0/\textit{92.7}}          & \multicolumn{1}{c|}{87.5/100}          & \multicolumn{1}{c|}{4.4/6.9}            & \multicolumn{1}{c|}{73.5/88.1}     & \multicolumn{1}{c|}{\textit{98.9}/\textit{99.8}}                                                  & \textit{99.5}/\textit{100} \\ \hline
\multicolumn{1}{|c|}{\textit{NV + SPC}}                                                                                              & \multicolumn{1}{c|}{41.6/62.1}     & \multicolumn{1}{c|}{82.0/97.3}          & \multicolumn{1}{c|}{41.0/61.0}          & \multicolumn{1}{c|}{\textit{66.5/\textbf{93.0}}} & \multicolumn{1}{c|}{79.0/90.0}          & \multicolumn{1}{c|}{65.6/84.4}         & \multicolumn{1}{c|}{8.2/12.8}           & \multicolumn{1}{c|}{\textbf{79.4}/\textbf{90.8}}     & \multicolumn{1}{c|}{\textbf{98.9}/\textbf{99.9}}                                         & 96.9/100          \\ \hline
\multicolumn{11}{|c|}{\textit{\textbf{MR-NetVLAD (Ours):}}}                                                                                                                                                                                                                                                                                                                                                                                                                                                                                                                                                                   \\ \hline
\multicolumn{1}{|c|}{\textit{BR}}                                                                                                    & \multicolumn{1}{c|}{\underline{\textbf{44.6}}/\underline{\textbf{67.2}}}     & \multicolumn{1}{c|}{59.5/88.3}         & \multicolumn{1}{c|}{\textit{73.3}/\textit{90.5}} & \multicolumn{1}{c|}{64.0/88.5}          & \multicolumn{1}{c|}{85.3/92.0}          & \multicolumn{1}{c|}{\underline{\textbf{93.8}}/\underline{\textbf{100}}}          & \multicolumn{1}{c|}{6.5/10.8}           & \multicolumn{1}{c|}{74.8/88.6}     & \multicolumn{1}{c|}{98.0/99.9}                                                  & \textbf{99.5}/\textbf{100}          \\ \hline
\multicolumn{1}{|c|}{\textit{BR + SPC}}                                                                                              & \multicolumn{1}{c|}{41.9/61.8}     & \multicolumn{1}{c|}{\textit{82.9}/\textit{98.2}} & \multicolumn{1}{c|}{40.5/62.4}          & \multicolumn{1}{c|}{\underline{\textbf{69.0}}/\underline{\textbf{93.5}}}          & \multicolumn{1}{c|}{82.3/90.7}          & \multicolumn{1}{c|}{68.8/90.6}         & \multicolumn{1}{c|}{\textit{14.8}/\textit{23.4}} & \multicolumn{1}{c|}{\underline{\textbf{80.6}}/\underline{\textbf{90.9}}}     & \multicolumn{1}{c|}{\underline{\textbf{98.9}}/\underline{\textbf{100}}}                                                   & \underline{\textbf{100}}/\underline{\textbf{100}}           \\ \hline
\multicolumn{1}{|c|}{\textit{BR + MLR}}                                                                                              & \multicolumn{1}{c|}{\textbf{44.3}/\textit{65.3}}     & \multicolumn{1}{c|}{69.4/94.6}         & \multicolumn{1}{c|}{71.0/89.1}          & \multicolumn{1}{c|}{\textbf{69.0}/\textit{90.5}}          & \multicolumn{1}{c|}{\textit{87.7}/\textbf{93.0}} & \multicolumn{1}{c|}{\textit{90.6}/\textbf{100}} & \multicolumn{1}{c|}{5.7/9.3}            & \multicolumn{1}{c|}{71.0/86.0}     & \multicolumn{1}{c|}{97.5/99.6}                                                  & 98.4/100          \\ \hline
\end{tabular}\end{table*}

\subsection{Scale-Specific vs Scale-Agnostic Clustering and Feature Aggregation}
Table~\ref{table:common_distrib_vocab} evaluates and compares the performance of $L=3$ MR-NetVLAD configuration (BR + MLR) while considering both the shared and scale-specific visual vocabulary. Particularly, the latter allows aggregation of scale-specific features $l \ \forall \{1,2,4\}$ by learning their respective scale-specific visual vocabulary. Here, the proportion of visual words per scale is set based on the amount of information available at that scale level, i.e., higher resolution has more words than the lower resolutions. Considering that, the distribution of scale-specific clusters configuration is set as $\{34,18,12\}$ and the scale-specific VLAD representations are generated independently per scale and concatenated before the final VLAD vector normalization. This leads to the same size VLAD vector as the our originally proposed implementation. It is evident from Table~\ref{table:common_distrib_vocab} that $L=3$ MR-NetVLAD with shared vocabulary outperforms its counterpart trained with a scale-specific vocabulary.

\renewcommand{\tabcolsep}{2.5pt}  
\begin{table}
\centering
\caption{Recall@N performance comparison with common and scale-specific vocabularies.}
\label{table:common_distrib_vocab}
\begin{tabular}{|c|c|c|c|c|c|}
\hline
\multirow{2}{*}{\textbf{\begin{tabular}[c]{@{}c@{}}MR-NetVLAD\\ (Setting)\end{tabular}}}           & \textbf{Datasets}                                                       & \textit{\begin{tabular}[c]{@{}c@{}}Pitts30k\\ Test\end{tabular}} & \textit{Kudamm} & \textit{\begin{tabular}[c]{@{}c@{}}Toyko24/7\\ Test\end{tabular}} & \textit{\begin{tabular}[c]{@{}c@{}}Oxford\\ SnVsOvr\end{tabular}} \\ \cline{2-6} 
                                                                                                   & \textbf{\begin{tabular}[c]{@{}c@{}}Vocabulary/\\ Clusters\end{tabular}} & R@1                                                              & R@1             & R@1                                                               & R@1                                                               \\ \hline
\multirow{2}{*}{\textit{\begin{tabular}[c]{@{}c@{}}L=3\\ l$\in$ \{1,2,4\}\\ (BR + MLR)\end{tabular}}} & Common/64                                                               & \textbf{86.4}                                                    & \textbf{41.1}   & \textbf{66.0}                                                     & \textbf{97.6}                                                     \\ \cline{2-6} 
                                                                                                   & \begin{tabular}[c]{@{}c@{}}Distributed/\\ \{34,18,12\}\end{tabular}   & 82.8                                                             & 32.5            & 47.9                                                              & 93.2                                                              \\ \hline
\end{tabular}
\vspace{-1mm}
\end{table}

\subsection{Cluster Distribution of Multi-Resolution Features}
Figure~\ref{figure:ms_clustr_contrubution} shows \textit{hard} assignment of multi-scale local features $P^l$ to different cluster centers. This is visualized both spatially at image level (left) and as histograms (right) for three different resolutions, that is, $L=3$ with $l \in \{1,2,4\}$. The contributions (bin counts) to a cluster from different resolutions are normalized per cluster center in the histogram. It can be observed that when viewing the same place with different image/feature resolutions, some of the clusters consistently capture the same semantic information across resolutions, e.g. see the maroon color which represents road in the image-level visualization. Yet, there are certain clusters which respond more to a particular resolution, e.g. orange color ($v=10$ in the histograms) for the lowest resolution in the last row. Overall, these feature distributions indicate complementarity across features from different low resolutions, which help enrich the overall representation.

\begin{figure}
	\centering
	\includegraphics[width=1.0\linewidth, height=0.75\linewidth]{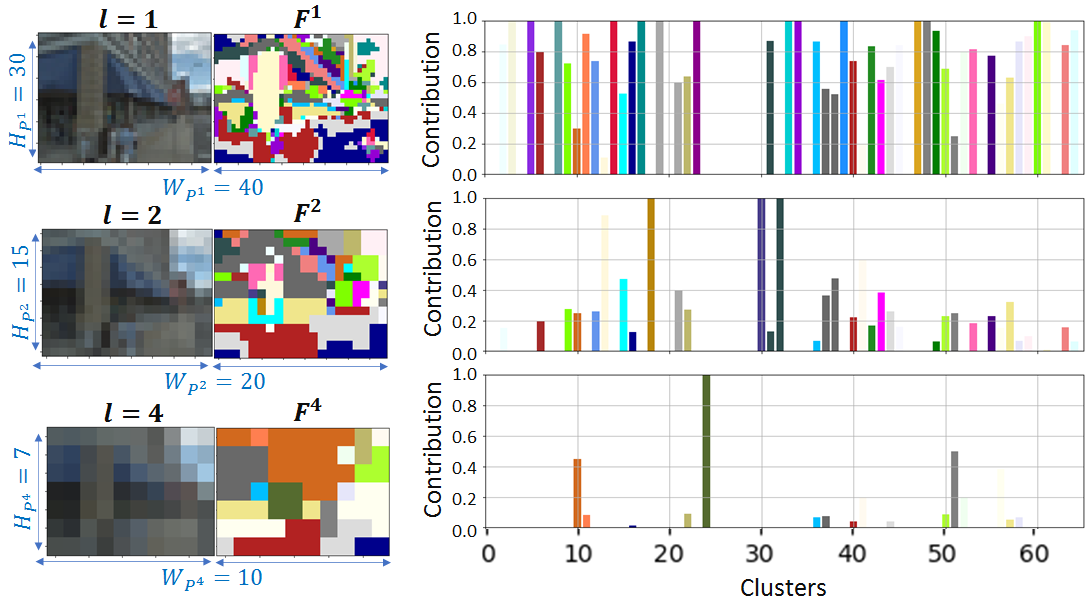}
	\caption{Cluster Distribution of Multi-Resolution Features at image level
(left) and as histograms (right).}
	\label{figure:ms_clustr_contrubution}
	\vspace{-5mm}
\end{figure}

\fi

\bibliographystyle{IEEEtran}
\bibliography{root}

\begin{thebibliography}{10}
\providecommand{\url}[1]{#1}
\csname url@samestyle\endcsname
\providecommand{\newblock}{\relax}
\providecommand{\bibinfo}[2]{#2}
\providecommand{\BIBentrySTDinterwordspacing}{\spaceskip=0pt\relax}
\providecommand{\BIBentryALTinterwordstretchfactor}{4}
\providecommand{\BIBentryALTinterwordspacing}{\spaceskip=\fontdimen2\font plus
\BIBentryALTinterwordstretchfactor\fontdimen3\font minus
  \fontdimen4\font\relax}
\providecommand{\BIBforeignlanguage}[2]{{%
\expandafter\ifx\csname l@#1\endcsname\relax
\typeout{** WARNING: IEEEtran.bst: No hyphenation pattern has been}%
\typeout{** loaded for the language `#1'. Using the pattern for}%
\typeout{** the default language instead.}%
\else
\language=\csname l@#1\endcsname
\fi
#2}}
\providecommand{\BIBdecl}{\relax}
\BIBdecl

\bibitem{lowry2015visual}
S.~Lowry, N.~S{\"u}nderhauf, P.~Newman, J.~J. Leonard, D.~Cox, P.~Corke, and
  M.~J. Milford, ``Visual place recognition: A survey,'' \emph{IEEE
  Transactions on Robotics}, vol.~32, no.~1, pp. 1--19, 2015.

\bibitem{garg2021your}
\BIBentryALTinterwordspacing
S.~Garg, T.~Fischer, and M.~Milford, ``Where is your place, visual place
  recognition?'' \emph{Proceedings of the Thirtieth International Joint
  Conference on Artificial Intelligence}, Aug 2021. [Online]. Available:
  \url{http://dx.doi.org/10.24963/ijcai.2021/603}
\BIBentrySTDinterwordspacing

\bibitem{masone2021survey}
C.~Masone and B.~Caputo, ``A survey on deep visual place recognition,''
  \emph{IEEE Access}, vol.~9, pp. 19\,516--19\,547, 2021.

\bibitem{cadena2016past}
C.~Cadena, L.~Carlone \emph{et~al.}, ``Past, present, and future of
  simultaneous localization and mapping: Toward the robust-perception age,''
  \emph{IEEE Transactions on Robotics}, vol.~32, no.~6, pp. 1309--1332, 2016.

\bibitem{Sarlin2021BackTT}
P.-E. Sarlin, A.~Unagar, M.~Larsson, H.~Germain, C.~Toft, V.~Larsson,
  M.~Pollefeys, V.~Lepetit, L.~Hammarstrand, F.~Kahl, and T.~Sattler, ``Back to
  the feature: Learning robust camera localization from pixels to pose,''
  \emph{2021 IEEE/CVF Conference on Computer Vision and Pattern Recognition
  (CVPR)}, pp. 3246--3256, 2021.

\bibitem{schonberger2016structure}
J.~L. Schonberger and J.-M. Frahm, ``Structure-from-motion revisited,'' in
  \emph{Proceedings of the IEEE conference on computer vision and pattern
  recognition}, 2016, pp. 4104--4113.

\bibitem{zhang2021visual}
X.~Zhang, L.~Wang, and Y.~Su, ``Visual place recognition: A survey from deep
  learning perspective,'' \emph{Pattern Recognition}, vol. 113, p. 107760,
  2021.

\bibitem{gabriele2021master}
G.~Trivigno, ``Deep learning for sequence-based visual geo-localization,''
  Torino TO, 2021.

\bibitem{doi:10.1177/0278364916679498}
\BIBentryALTinterwordspacing
W.~Maddern, G.~Pascoe, C.~Linegar, and P.~Newman, ``1 year, 1000 km: The oxford
  robotcar dataset,'' \emph{The International Journal of Robotics Research},
  vol.~36, no.~1, pp. 3--15, 2017. [Online]. Available:
  \url{https://doi.org/10.1177/0278364916679498}
\BIBentrySTDinterwordspacing

\bibitem{arandjelovic2016netvlad}
R.~Arandjelovic, P.~Gronat, A.~Torii, T.~Pajdla, and J.~Sivic, ``Netvlad: Cnn
  architecture for weakly supervised place recognition,'' in \emph{Proceedings
  of the IEEE conference on computer vision and pattern recognition}, 2016, pp.
  5297--5307.

\bibitem{chen2018learning}
Z.~Chen, L.~Liu, I.~Sa, Z.~Ge, and M.~Chli, ``Learning context flexible
  attention model for long-term visual place recognition,'' \emph{IEEE Robotics
  and Automation Letters}, vol.~3, no.~4, pp. 4015--4022, 2018.

\bibitem{hausler2021patch}
S.~Hausler, S.~Garg, M.~Xu, M.~Milford, and T.~Fischer, ``Patch-netvlad:
  Multi-scale fusion of locally-global descriptors for place recognition,'' in
  \emph{Proceedings of the IEEE/CVF Conference on Computer Vision and Pattern
  Recognition}, 2021, pp. 14\,141--14\,152.

\bibitem{8461081}
Y.~Latif, R.~Garg, M.~Milford, and I.~Reid, ``Addressing challenging place
  recognition tasks using generative adversarial networks,'' in \emph{2018 IEEE
  International Conference on Robotics and Automation (ICRA)}, 2018, pp.
  2349--2355.

\bibitem{garg2018lost}
S.~Garg, N.~Suenderhauf, and M.~Milford, ``Lost? appearance-invariant place
  recognition for opposite viewpoints using visual semantics,''
  \emph{Proceedings of Robotics: Science and Systems XIV}, 2018.

\bibitem{7989305}
T.~Naseer, G.~L. Oliveira, T.~Brox, and W.~Burgard, ``Semantics-aware visual
  localization under challenging perceptual conditions,'' in \emph{2017 IEEE
  International Conference on Robotics and Automation (ICRA)}, 2017, pp.
  2614--2620.

\bibitem{DBLP:journals/ftrob/GargSDMCCWCRGCM20}
\BIBentryALTinterwordspacing
S.~Garg, N.~S{\"{u}}nderhauf, F.~Dayoub, D.~Morrison, A.~Cosgun, G.~Carneiro,
  Q.~Wu, T.~Chin, I.~D. Reid, S.~Gould, P.~Corke, and M.~Milford, ``Semantics
  for robotic mapping, perception and interaction: {A} survey,'' \emph{Found.
  Trends Robotics}, vol.~8, no. 1-2, pp. 1--224, 2020. [Online]. Available:
  \url{https://doi.org/10.1561/2300000059}
\BIBentrySTDinterwordspacing

\bibitem{6594910}
F.~Endres, J.~Hess, J.~Sturm, D.~Cremers, and W.~Burgard, ``3-d mapping with an
  rgb-d camera,'' \emph{IEEE Transactions on Robotics}, vol.~30, no.~1, pp.
  177--187, 2014.

\bibitem{pepperell2014all}
E.~Pepperell, P.~I. Corke, and M.~J. Milford, ``All-environment visual place
  recognition with smart,'' in \emph{IEEE International Conference on Robotics
  and Automation}, 2014, pp. 1612--1618.

\bibitem{7989618}
R.~Dubé, D.~Dugas, E.~Stumm, J.~Nieto, R.~Siegwart, and C.~Cadena, ``Segmatch:
  Segment based place recognition in 3d point clouds,'' in \emph{2017 IEEE
  International Conference on Robotics and Automation (ICRA)}, 2017, pp.
  5266--5272.

\bibitem{8593562}
P.~Yin, L.~Xu, Z.~Liu, L.~Li, H.~Salman, Y.~He, W.~Xu, H.~Wang, and H.~Choset,
  ``Stabilize an unsupervised feature learning for lidar-based place
  recognition,'' in \emph{2018 IEEE/RSJ International Conference on Intelligent
  Robots and Systems (IROS)}, 2018, pp. 1162--1167.

\bibitem{milford2012seqslam}
M.~J. Milford and G.~F. Wyeth, ``Seqslam: Visual route-based navigation for
  sunny summer days and stormy winter nights,'' in \emph{2012 IEEE
  international conference on robotics and automation}.\hskip 1em plus 0.5em
  minus 0.4em\relax IEEE, 2012, pp. 1643--1649.

\bibitem{garg2021seqmatchnet}
S.~Garg, M.~Vankadari, and M.~Milford, ``Seqmatchnet: Contrastive learning with
  sequence matching for place recognition \& relocalization,'' in \emph{5th
  Annual Conference on Robot Learning}, 2021.

\bibitem{Garg2021SeqNetLD}
S.~Garg and M.~Milford, ``Seqnet: Learning descriptors for sequence-based
  hierarchical place recognition,'' \emph{IEEE Robotics and Automation
  Letters}, vol.~6, pp. 4305--4312, 2021.

\bibitem{neubert2021vector}
P.~Neubert, S.~Schubert, K.~Schlegel, and P.~Protzel, ``Vector semantic
  representations as descriptors for visual place recognition,'' 2021.

\bibitem{schubert2021fast}
S.~Schubert, P.~Neubert, and P.~Protzel, ``Fast and memory efficient graph
  optimization via icm for visual place recognition,'' 2021.

\bibitem{8700608}
J.~Yu, C.~Zhu, J.~Zhang, Q.~Huang, and D.~Tao, ``Spatial pyramid-enhanced
  netvlad with weighted triplet loss for place recognition,'' \emph{IEEE
  Transactions on Neural Networks and Learning Systems}, vol.~31, no.~2, pp.
  661--674, 2020.

\bibitem{8382272}
F.~Radenović, G.~Tolias, and O.~Chum, ``Fine-tuning cnn image retrieval with
  no human annotation,'' \emph{IEEE Transactions on Pattern Analysis and
  Machine Intelligence}, vol.~41, no.~7, pp. 1655--1668, 2019.

\bibitem{le2020city}
\BIBentryALTinterwordspacing
D.~C. Le and C.~Youn, ``City-scale visual place recognition with deep local
  features based on multi-scale ordered {VLAD} pooling,'' \emph{CoRR}, vol.
  abs/2009.09255, 2020. [Online]. Available:
  \url{https://arxiv.org/abs/2009.09255}
\BIBentrySTDinterwordspacing

\bibitem{8961602}
Z.~Li, A.~Zhou, M.~Wang, and Y.~Shen, ``Deep fusion of multi-layers salient cnn
  features and similarity network for robust visual place recognition,'' in
  \emph{2019 IEEE International Conference on Robotics and Biomimetics
  (ROBIO)}.\hskip 1em plus 0.5em minus 0.4em\relax IEEE, 2019, pp. 22--29.

\bibitem{xin2019localizing}
Z.~Xin, Y.~Cai, T.~Lu, X.~Xing, S.~Cai, J.~Zhang, Y.~Yang, and Y.~Wang,
  ``Localizing discriminative visual landmarks for place recognition,'' IEEE,
  pp. 5979--5985, 2019.

\bibitem{mao2019learning}
J.~Mao, X.~Hu, X.~He, L.~Zhang, L.~Wu, and M.~J. Milford, ``Learning to fuse
  multiscale features for visual place recognition,'' \emph{IEEE Access},
  vol.~7, pp. 5723--5735, 2019.

\bibitem{sunderhauf2013we}
N.~S{\"u}nderhauf, P.~Neubert, and P.~Protzel, ``Are we there yet? challenging
  seqslam on a 3000 km journey across all four seasons,'' in \emph{Proc. of
  workshop on long-term autonomy, IEEE international conference on robotics and
  automation (ICRA)}, 2013, pp. 1--3.

\bibitem{chen2017deep}
Z.~Chen, A.~Jacobson, N.~S{\"u}nderhauf, B.~Upcroft, L.~Liu, C.~Shen, I.~Reid,
  and M.~Milford, ``Deep learning features at scale for visual place
  recognition,'' in \emph{IEEE International Conference on Robotics and
  Automation}, 2017, pp. 3223--3230.

\bibitem{zaffar2020memorable}
M.~Zaffar, S.~Ehsan, M.~Milford, and K.~D. McDonald-Maier, ``Memorable maps: A
  framework for re-defining places in visual place recognition,'' \emph{IEEE
  Transactions on Intelligent Transportation Systems}, 2020.

\bibitem{zaffar2020vpr}
M.~Zaffar, S.~Garg, M.~Milford, J.~Kooij, D.~Flynn, K.~McDonald-Maier, and
  S.~Ehsan, ``Vpr-bench: An open-source visual place recognition evaluation
  framework with quantifiable viewpoint and appearance change,''
  \emph{International Journal of Computer Vision}, pp. 1--39, 2021.

\bibitem{filliat2007visual}
D.~Filliat, ``A visual bag of words method for interactive qualitative
  localization and mapping,'' in \emph{IEEE International Conference on
  Robotics and Automation}, 2007, pp. 3921--3926.

\bibitem{cummins2011appearance}
M.~Cummins and P.~Newman, ``Appearance-only {SLAM} at large scale with
  {FAB-MAP} 2.0,'' \emph{International Journal of Robotics Research}, vol.~30,
  no.~9, pp. 1100--1123, 2011.

\bibitem{lowe2004distinctive}
D.~G. Lowe, ``Distinctive image features from scale-invariant keypoints,''
  \emph{International Journal of Computer Vision}, vol.~60, no.~2, pp. 91--110,
  2004.

\bibitem{bay2006surf}
H.~Bay, T.~Tuytelaars \emph{et~al.}, ``Surf: Speeded up robust features,'' in
  \emph{Proc. European Conference on Computer Vision}, 2006, pp. 404--417.

\bibitem{8972582}
M.~{Zaffar}, S.~{Ehsan}, M.~J. {Milford}, and K.~{McDonald-Maier}, ``Cohog: A
  light-weight, compute-efficient and training-free visual place recognition
  technique for changing environments,'' \emph{IEEE Robotics and Automation
  Letters}, pp. 1--1, 2020.

\bibitem{dalal2005histograms}
N.~Dalal \emph{et~al.}, ``Histograms of oriented gradients for human
  detection,'' in \emph{CVPR}, vol.~1.\hskip 1em plus 0.5em minus 0.4em\relax
  IEEE Computer Society, 2005, pp. 886--893.

\bibitem{oliva2006building}
A.~Oliva and A.~Torralba, ``Building the gist of a scene: The role of global
  image features in recognition,'' \emph{Progress in brain research}, vol. 155,
  pp. 23--36, 2006.

\bibitem{babenko2015aggregating}
A.~Babenko and V.~Lempitsky, ``Aggregating local deep features for image
  retrieval,'' in \emph{IEEE International Conference on Computer Vision},
  2015, pp. 1269--1277.

\bibitem{jegou2010aggregating}
H.~J{\'e}gou, M.~Douze, C.~Schmid, and P.~P{\'e}rez, ``Aggregating local
  descriptors into a compact image representation,'' in \emph{IEEE Conference
  on Computer Vision and Pattern Recognition}, 2010, pp. 3304--3311.

\bibitem{7301270}
H.~Azizpour, A.~S. Razavian, J.~Sullivan, A.~Maki, and S.~Carlsson, ``From
  generic to specific deep representations for visual recognition,'' in
  \emph{2015 IEEE Conference on Computer Vision and Pattern Recognition
  Workshops (CVPRW)}, 2015, pp. 36--45.

\bibitem{Tolias2015ParticularOR}
G.~Tolias, R.~Sicre, and H.~J{\'e}gou, ``Particular object retrieval with
  integral max-pooling of cnn activations,'' \emph{International Conference on
  Learning Representations}, vol. abs/1511.05879, 2015.

\bibitem{chen2017only}
Z.~Chen, F.~Maffra, I.~Sa, and M.~Chli, ``Only look once, mining distinctive
  landmarks from convnet for visual place recognition,'' in \emph{2017 IEEE/RSJ
  International Conference on Intelligent Robots and Systems (IROS)}.\hskip 1em
  plus 0.5em minus 0.4em\relax IEEE, 2017, pp. 9--16.

\bibitem{cao2020unifying}
B.~Cao, A.~F. de~Ara{\'u}jo, and J.~Sim, ``Unifying deep local and global
  features for image search,'' in \emph{ECCV}, 2020.

\bibitem{revaud2019learning}
J.~Revaud, J.~Almaz{\'a}n, R.~S. de~Rezende, and C.~R. de~Souza, ``Learning
  with average precision: Training image retrieval with a listwise loss,''
  \emph{2019 IEEE/CVF International Conference on Computer Vision (ICCV)}, pp.
  5106--5115, 2019.

\bibitem{zaffar2019levelling}
M.~Zaffar, A.~Khaliq, S.~Ehsan \emph{et~al.}, ``Levelling the playing field: A
  comprehensive comparison of visual place recognition approaches under
  changing conditions,'' \emph{arXiv preprint arXiv:1903.09107}, 2019.

\bibitem{gordo2017end}
A.~Gordo \emph{et~al.}, ``End-to-end learning of deep visual representations
  for image retrieval,'' \emph{International Journal of Computer Vision}, vol.
  124, no.~2, pp. 237--254, 2017.

\bibitem{chen2016attention}
L.-C. Chen, Y.~Yang, J.~Wang, W.~Xu \emph{et~al.}, ``Attention to scale:
  Scale-aware semantic image segmentation,'' in \emph{Proc. IEEE conference on
  Computer Vision and Pattern Recognition}, 2016, pp. 3640--3649.

\bibitem{khaliq2018holistic}
A.~{Khaliq}, S.~{Ehsan}, Z.~{Chen}, M.~{Milford}, and otherss, ``A holistic
  visual place recognition approach using lightweight cnns for significant
  viewpoint and appearance changes,'' \emph{IEEE Transactions on Robotics}, pp.
  1--9, 2019.

\bibitem{Xin2019RealTimeVP}
Z.~Xin, X.~Cui, J.~Zhang, Y.~Yang, and Y.~Wang, ``Real-time visual place
  recognition based on analyzing distribution of multi-scale cnn landmarks,''
  \emph{Journal of Intelligent \& Robotic Systems}, vol.~94, pp. 777--792,
  2019.

\bibitem{Zhu_2018}
Y.~Zhu, J.~Wang, L.~Xie, and L.~Zheng, ``Attention-based pyramid aggregation
  network for visual place recognition,'' \emph{Proceedings of the 26th ACM
  international conference on Multimedia}, pp. 99--107, 2018.

\bibitem{farandjelovic2013all}
R.~Arandjelovic and A.~Zisserman, ``All about {VLAD},'' in \emph{IEEE
  Conference on Computer Vision and Pattern Recognition}, 2013, pp. 1578--1585.

\bibitem{torii2013visual}
A.~Torii, J.~Sivic, T.~Pajdla, and M.~Okutomi, ``Visual place recognition with
  repetitive structures,'' in \emph{CVPR}, 2013, pp. 883--890.

\bibitem{torii201524}
A.~Torii, R.~Arandjelovic, J.~Sivic, M.~Okutomi, and T.~Pajdla, ``24/7 place
  recognition by view synthesis,'' in \emph{IEEE Conference on Computer Vision
  and Pattern Recognition}, 2015, pp. 1808--1817.

\bibitem{10.1007/978-3-319-49409-8_74}
E.~Sizikova, V.~K. Singh, B.~Georgescu, M.~Halber, K.~Ma, and T.~Chen,
  ``Enhancing place recognition using joint intensity - depth analysis and
  synthetic data,'' in \emph{Computer Vision -- ECCV 2016 Workshops}, G.~Hua
  and H.~J{\'e}gou, Eds.\hskip 1em plus 0.5em minus 0.4em\relax Cham: Springer
  International Publishing, 2016, pp. 901--908.

\bibitem{simonyan2014very}
K.~Simonyan \emph{et~al.}, ``Very deep convolutional networks for large-scale
  image recognition,'' \emph{International Conference on Learning
  Representations}, 2015.

\bibitem{peng2021semantic}
G.~Peng, Y.~Yue, J.~Zhang, Z.~Wu, X.~Tang, and D.~Wang, ``Semantic reinforced
  attention learning for visual place recognition,'' in \emph{2021 IEEE
  International Conference on Robotics and Automation (ICRA)}.\hskip 1em plus
  0.5em minus 0.4em\relax IEEE, 2021, pp. 13\,415--13\,422.

\bibitem{jin2017learned}
H.~J. Kim, E.~Dunn \emph{et~al.}, ``Learned contextual feature reweighting for
  image geo-localization,'' in \emph{IEEE Conference on Computer Vision and
  Pattern Recognition}, 2017, pp. 2136--2145.

\bibitem{lindeberg1994scale}
T.~Lindeberg, ``Scale-space theory: A basic tool for analyzing structures at
  different scales,'' \emph{Journal of applied statistics}, vol.~21, no. 1-2,
  pp. 225--270, 1994.

\bibitem{burt1987laplacian}
P.~J. Burt and E.~H. Adelson, ``The laplacian pyramid as a compact image
  code,'' in \emph{Readings in computer vision}.\hskip 1em plus 0.5em minus
  0.4em\relax Elsevier, 1987, pp. 671--679.

\bibitem{crowley2002fast}
J.~L. Crowley, O.~Riff, and J.~H. Piater, ``Fast computation of characteristic
  scale using a half octave pyramid,'' in \emph{International Conference on
  Scale-Space Theories in Computer Vision}, 2002.

\end{thebibliography}


\end{document}